\documentclass[a4paper]{article}[12pts]

\usepackage[english]{babel}
\usepackage[utf8x]{inputenc}
\usepackage[T1]{fontenc}

\normalsize

\usepackage[a4paper,top=1in,bottom=1in,left=1in,right=1in,marginparwidth=1in]{geometry}

\usepackage{natbib}
\usepackage{framed}
\usepackage{setspace}
\usepackage{amsmath}
\usepackage{amssymb} 
\usepackage{amsthm}
\usepackage{amsfonts, mathtools}
\usepackage{bm}
\usepackage{bbm}
\usepackage{comment}
\usepackage{graphicx}
\usepackage{multirow, booktabs}
\usepackage{caption}
\usepackage{subcaption}
\usepackage{algorithm,algpseudocode}
\usepackage[colorinlistoftodos]{todonotes}
\usepackage{hyperref}
\hypersetup{colorlinks=true, allcolors=blue}

\usepackage{textgreek}
\usepackage{adjustbox}
\usepackage{mdframed}

\definecolor{azure}{rgb}{0.9, 0.95, 1.0}

\theoremstyle{plain}
\newtheorem{theorem}{Theorem}[section]
\newtheorem{proposition}[theorem]{Proposition}
\newtheorem{lemma}[theorem]{Lemma}
\newtheorem{corollary}[theorem]{Corollary}
\theoremstyle{definition}
\newtheorem{definition}[theorem]{Definition}
\newtheorem{assumption}[theorem]{Assumption}

\theoremstyle{remark}

\usepackage[table]{xcolor} 
\usepackage{makecell}

\definecolor{sotagray}{gray}{0.9}

\usepackage[most]{tcolorbox}

\newtcolorbox{templatebox}{
  colback=black!5,
  colframe=black!75,
  boxrule=0.5pt,
  sharp corners,
  fontupper=\small\ttfamily,
  breakable
}

\newtcolorbox{promptbox}{
  colback=blue!5!white,
  colframe=blue!60!black,
  boxrule=0.5pt,
  sharp corners,
  fontupper=\small,
  breakable
}

\usepackage{tikz}
\usetikzlibrary{shapes.geometric, arrows.meta, positioning, fit} 

\renewcommand{\thefootnote}{\fnsymbol{footnote}}

\newcommand{\Alg}[1]{\mathrm{Alg}\,#1}
\newcommand{\FalgInf}[2]{\mathcal{F}_{\Alg{#1},\infty}\!\big(#2\big)}
\newcommand{\Falgp}[2]{\mathcal{F}_{\Alg{#1},p}\!\big(#2\big)}

\title{Exploration-free Algorithms for Multi-group Mean Estimation}
\author{Ziyi Wei$^\dagger$, \ Huaiyang Zhong$^\dagger$, \ Xiaocheng Li$^\ddagger$}
\date{\small
$^\dagger$ Grado Department of Industrial and Systems Engineering, Virginia Tech\\
$^\ddagger$
Imperial College Business School, Imperial College London
}

\begin{document}
\maketitle
\onehalfspacing

\def\thefootnote{}\relax\footnotetext{Corresponding to  Ziyi Wei (\url{ziyiw@vt.edu}). 
}
\begin{abstract}
We address the problem of multi-group mean estimation, which seeks to allocate a finite sampling budget across multiple groups to obtain uniformly accurate estimates of their means. Unlike classical multi-armed bandits, whose objective is to minimize regret by identifying and exploiting the best arm, the optimal allocation in this setting requires sampling every group on the order of $\Theta(T)$ times. This fundamental distinction makes exploration-free algorithms both natural and effective. Our work makes three contributions. First, we strengthen the existing results on subgaussian variance concentration using the Hanson-Wright inequality and identify a class of strictly subgaussian distributions that yield sharper guarantees. Second, we design exploration-free non-adaptive and adaptive algorithms, and we establish tighter regret bounds than the existing results. Third, we extend the framework to contextual bandit settings, an underexplored direction, and propose algorithms that leverage side information with provable guarantees. Overall, these results position exploration-free allocation as a principled and efficient approach to multi-group mean estimation, with potential applications in experimental design, personalization, and other domains requiring accurate multi-group inference.
\end{abstract}

\section{Introduction}
We study the problem of multi-group mean estimation, where the task is to allocate a limited sampling budget across multiple groups in order to estimate their means uniformly well. This problem arises naturally in polling, survey design, marketing, and other settings where representative estimates across diverse groups are required. A key feature distinguishing this setting from classical reward--maximization bandits is that the optimal allocation requires sampling every arm on the order of $\Theta(T)$ times, rather than focusing as much as possible on the best option. This structural property suggests that explicit exploration phases are unnecessary and opens the door to exploration-free algorithms.  

Contextual information makes the problem even more relevant in real-world applications such as healthcare \citep{bastani2020online, du2024contextual}, recommendation systems \citep{agarwal2009online, li2010contextual}, and dynamic pricing \citep{qiang2016dynamic, ban2021personalized}, where side information fundamentally shapes the reward distributions and motivates the estimation of context-dependent group parameters. Accurate estimation in this richer setting is crucial for interpretable personalization, robust policy design, and fairness considerations. 

\paragraph{Literature Review.}
In the traditional bandit setting, several groups of papers have studied this problem and proposed several extensions on the objective metrc \citep{antos2008active,antos2010active,carpentier2011upper,shekhar20a,aznag2023an}. 
For linear models, in particular linear and contextual bandits, the central task reduces to estimating the unknown coefficient vector $\beta$. 
\citet{riquelme2017active} analyzed this problem in the multi--linear regression setting under the assumption that both the noise and the context vectors are Gaussian. \citet{fontaine2021online} focused on the univariate case of $\beta^{*}$ and evaluated performance using the metric $\mathbb{E}[\|\hat{\beta}-\beta^{*}\|^{2}]$, their analysis accommodates heterogeneous subgaussian noise while assuming that the context vectors have unit norm. Beyond these closely related works, the literature on group mean estimation also connects to research in areas such as best-arm identification, conservative bandits, and experimental design. We defer more discussions and other related literature to Appendix~\ref{appendix: literature}. 

\paragraph{Contributions.}
Our contributions are threefold: First, we strengthen the existing results on subgaussian variance concentration using the Hanson-Wright inequality and identify a class of strictly subgaussian distributions that yield sharper guarantees. Second, we design non-adaptive and adaptive algorithms that are exploration-free, and we establish tighter regret bounds than the existing results. Third, we extend the framework to contextual bandit settings, an underexplored direction, and propose algorithms that leverage side information with provable guarantees. Theoretically, our results reveal certain structural properties of the problem, such as exploration-free design, the failure mode of UCB-type algorithms for the contextual setting, and also connections with the best-arm identification problem.

\section{Problem Setup}
In this section, we present the problem setup of \textit{multi-group mean estimation}. Consider $K$ alternatives (also known as arms in the multi-armed bandits literature), each of which is associated with a random outcome/reward. Specifically, the outcome of $k$-th alternative follows a distribution $\mathcal{P}_k$ with unknown mean $\mu_k$ and variance $\sigma_k^2.$ We consider an online learning setup where there is a finite horizon $T$. At each time $t=1,...,T$, the decision maker chooses one alternative $k_t \in \{1,...,K\}$, and then (s)he observes and only observes the outcome of a realization $X_{k_t}\sim\mathcal{P}_k.$ Notation-wise, if we take the standpoint of each alternative $k$, we use $X_{k,t}$ to denote the $t$-th observations we collect from the $k$-th alternative. Accordingly, we use $n_{k}$ to denote the total number of times the $k$-th alternative is chosen throughout the horizon $T$. In this way, we can estimate the mean of alternative $k$ by the average of the observations
\begin{align*}
    \hat{\mu}_{k}(T) \coloneqq \frac{1}{n_{k}} \sum_{t=1}^{n_{k}} X_{k,t}.
\end{align*}

The task of \textit{multi-group mean estimation} aims to accurately estimate the mean for all alternatives simultaneously. By the end of the horizon, the estimation error for alternative $k$ is
\begin{align}\label{eq:MAB_objective}
\mathbb{E}_{\mathcal{P}_{k}} [(\hat{\mu}_{k}(T)-\mu_{k})^{2}] = \frac{\sigma_{k}^{2}}{n_{k}}.
\end{align}
To aggregate the errors across all the alternatives, 
\citet{aznag2023an} propose the following objective for multi-group mean estimation, for $p>0$,
\begin{align*}
    R_{p}(\boldsymbol{n}) \coloneqq \left\| \left\{ \frac{\sigma^{2}_{k}}{n_{k}} \right\}_{k=1}^{K} \right\|_{p} 
    =
    \begin{cases}
        ( \sum_{k=1}^{K} \frac{\sigma_k^{2p}}{n_k^{p}})^{\frac{1}{p}}, & \text{if } p < \infty, \\
        \max\limits_{1 \leq k \leq K} \frac{\sigma_k^{2}}{n_k}, & \text{if } p = \infty.
    \end{cases}
\end{align*}
We note the objective only involves $\bm{n}=(n_1,...,n_K)$, the number of observations for each alternative, as the decision variables. Intuitively, the objective requires budgeting the observations in a way that we obtain a uniformly good mean estimation for all the alternatives. Importantly, the variance $\sigma_k^2$ is unknown and has to be estimated from the observations as well.

\begin{proposition}
Suppose one knows $\sigma_k^2$, then the optimal allocation $\bm{n}^*=(n_1^*,...,n_{K}^*)$ is given by
$$n_{k}^* = \frac{\sigma_{k}^{q}}{\sum_{j=1}^{K} \sigma_{j}^{q}} \cdot T$$
where $q\coloneqq \tfrac{2p}{p+1}$ if $p$ is finite and $q=2$ if $p =\infty$ and the optimal objective value is:
\begin{align*}
R_{p}(\bm{n}^{*}) = \frac{1}{T}\left(\sum_{k=1}^{K} \sigma_{k}^{q}\right)^{\frac{2}{q}}.
\end{align*}
\label{prop:optimal}
\end{proposition}

Proposition \ref{prop:optimal} is obtained from solving the optimization problem with the knowledge of $\sigma^2_k$. Without the knowledge, an algorithm can never beat this objective value, and will be benchmarked against the value to measure the algorithm's performance. We also note that the optimal solution depends both on $\sigma_k^2$'s and the norm $p$, and the optimal value scales on the order of $1/T.$

Before we talk about our algorithms, we first introduce some assumptions and basic inequalities to help our analyses. First, we introduce the concept of subgaussian and strictly subgaussian. 
\begin{definition}\label{def:subgaussian}
A random variable $X$ with distribution $\mathcal{P}_{X}$ is $\sigma$--subgaussian such that for all $t \in \mathbb{R}$:
    \begin{align}\label{eq:subgaussian_definition}
     \mathbb{E}_{X \sim \mathcal{P}_{X}}\left[ \exp(tX) \right] \leq \exp(t^{2}\sigma^{2}/2). 
    \end{align}
Moreover, if $X$'s variance is $\sigma_X^2$, we say $X$ is strictly-subgaussian if it satisfies (\ref{eq:subgaussian_definition}) with \( \sigma^2  =  \sigma_{X}^{2}\).
\end{definition}

\begin{assumption}\label{assump:subgaussian}
Throughout our paper, we assume
\begin{itemize}
\item[(a)] $\mathcal{P}_{k}$ follows $\sigma$--subgaussian for all $k$.
\item[(b)] \(\sigma^{2}_{\min} = \min_{k=1,...,K}\sigma^{2}_{k}\) is positive. 
\end{itemize}
We also assume $\sigma$ is known.
\end{assumption}

These are two mild assumptions: Part (a) is commonly assumed in multi-armed bandits literature, and Part (b) simply says that all the alternatives are random (if there is a deterministic one, we don't really need to estimate its mean).

\subsection{Subgaussian variance concentration}

While the optimal allocation scheme is determined by $\sigma_k^2$ as in Proposition \ref{prop:optimal}, any algorithm that solves the problem should naturally involve some variance estimation, i.e., estimating $\sigma_k^2$ from observations. So we first state several concentration inequalities related to variance estimation. For this subsection, we state the results for a general random variable $X$. For $n$ i.i.d. observations $X_1,...,X_n$, one can construct a variance estimator 
\begin{align}
    \hat{\sigma}_{n}^{2} \coloneqq \frac{1}{n-1}\sum_{i=1}^{n}(X_{i}-\bar{X})^2
\label{eqn:var_estimate}
\end{align}
where $\bar{X}$ is the sample mean. Let $\sigma_X^2$ be the true variance of $X$. 

\begin{lemma} \label{lemma:variance_concentrarion}
Suppose $X$ is $\sigma$-subgaussian, then we have
\begin{align*}
    \mathbb{P}\left(|\hat\sigma_n^{2}-\sigma_X^{2}|\ge 4\sigma^{2}f(n)\sqrt{\frac{2\log(1/\delta)}{n-1}}+\frac{6\sigma^{2}\log(1/\delta)}{n}\right)\le2\delta,
\end{align*}
where $f(n) = (1+\sqrt{n-1})/\sqrt{n}$. Specifically, if $X$ is strictly--subgaussian, then $f(n) = (1+\sqrt{(n-1)/8})/\sqrt{n}$, and $\sigma$ can be replaced with $\sigma_{X}$. 
\end{lemma}

For the results, we refer to Appendix~\ref{appendix: subgaussian_subgamma} for the proof sketch and a detailed discussion. It utilizes the structure of the variance estimator and gets rid of the term $\sqrt{\log (1/\delta)}$ used in the previous work \citep{aznag2023an}. The lemma tells that the error of the variance estimator shrinks at a rate of $1/{\sqrt{n}}$. A subtle point is that when $X$ is strictly subgaussian, the constants are improved and the subgaussian parameter $\sigma$ is improved to the true variance. For the case of the Gaussian distribution, the bound can be further tightened based on Lemma~1 in \citet{laurent2000adaptive}:
\begin{align}\label{eq:gaussian_concentration}
    \mathbb{P}\big(\hat{\sigma}_{n}^{2}-\sigma_{X}^{2} 
    \geq 2\sigma_{X}^{2}\sqrt{\frac{\log(1/\delta)}{n-1}} + \frac{2\sigma^{2}_{X}\log(1/\delta)}{n-1} \big) \leq \delta, \quad \mathbb{P}\big(\sigma_{X}^{2} - \hat{\sigma}_{n}^{2} 
    \geq 2\sigma_{X}^{2}\sqrt{\frac{\log(1/\delta)}{n-1}} \big) \leq \delta.
\end{align}
For these bounds, the constants of the leading-order term are close to those of the strictly subgaussian case, which suggests that the strictly subgaussian structure offers a level of tail control comparable to the most ideal Gaussian case.

\section{Non-adaptive-Style Algorithm}\label{section3}

We first present a non-adaptive algorithm for the problem. The algorithm doesn't require knowing $\sigma_k^2$ exactly, but requires a knowledge of a lower bound $\underline{\sigma}^2$, i.e. for all $k=1,...,K,$
\begin{align*}
    \sigma_{k}^{2} \geq \underline{\sigma}^2 > 0.
\end{align*}
The knowledge of $\underline{\sigma}^2$ can usually be obtained from historical data or domain knowledge. We will fully remove this requirement in the next section. Here we use the setup to generate more intuitions for the algorithm design. It also provides insights into the special structure of multi-group mean estimation and how it differs from multi-armed bandits and best arm identification \citep{audibert2010best}.

\begin{algorithm}[ht!]
\renewcommand{\thealgorithm}{1}
\caption{Non-adaptive allocation}
\begin{algorithmic}[1]
\Require \( T \), initial length \( \tau \),  constant \( q \)

\State \textbf{Phase 1: Uniformly select and estimate}
\For{each alternative \( k = 1, \ldots, K \)}
    \State Choose \( k \) for \( \tau \) rounds (time periods)
\EndFor
\State Estimate $\hat{\sigma}_{k,\tau}^{2}$ with Equation~\eqref{eqn:var_estimate}.

\State \textbf{Phase 2: Allocate the remaining periods}
    \State  Compute allocation weight:
    \begin{align}\label{eq:allocation_lambda}
        \lambda_{k,\tau}  = \frac{\hat{\sigma}_{k,\tau}^{q}}{\sum_{j=1}^{K} \hat{\sigma}_{j,\tau}^{q}}        
    \end{align}

\For{each alternative \( k = 1, \ldots, K \)}
    \State Choose \( k \) for \( \lambda_{k,\tau} T - \tau \) rounds
    \State Calculate: $$\hat{\mu}_{k}(T) = \frac{1}{\lambda_{k,\tau}T}\sum_{t=1}^{\lambda_{k,\tau}T} X_{k,t}$$
\EndFor
\State \textbf{Output:} Final estimates \( \{\hat{\mu}_{k}(T)\}_{k=1}^{K} \)
\end{algorithmic}
\label{alg:FETE}
\end{algorithm}

Algorithm \ref{alg:FETE} requires two inputs (in addition to the horizon length $T$): the initial length $\tau$ and the constant $q$. First, $q$ is determined by the norm $p$ in the performance measure, and the definition is given in Proposition \ref{prop:optimal}. Second, the exploration length $\tau$ is given by the following
\begin{align}\label{eq:tau_nonadaptive}
    \tau \coloneqq \frac{\underline{\sigma}^q}{\underline{\sigma}^q+(K-1)\cdot \sigma^q}\cdot T
\end{align} 
where $\underline{\sigma}^2$ is the variance lower bound and $\sigma$ is the subgaussian parameter. It is easy to verify (from Proposition \ref{prop:optimal}) that such a choice ensures that $$\tau \le \min_{k=1,...,K} n_k^*.$$ 
For notation simplicity, we just assume all the values are integers and omit the floor symbol. The algorithm is a direct implication of Proposition \ref{prop:optimal}. Recall that Proposition \ref{prop:optimal} says the optimal allocation scheme $\bm{n}^*$ depends on the true $\sigma_k^2.$ The algorithm basically estimates the variances with observations in the exploration and then allocates the remaining time periods according to the optimal solution structure in Proposition \ref{prop:optimal}. We call the algorithm as \textit{non-adaptive} allocation in that the variances are estimated just based on the initial $\tau$ observations, and then the allocation scheme is determined accordingly and will not be adaptively adjusted later. This non-adaptive nature of the algorithm resembles the non-adaptive design \citep{glynn2004large} for the best-arm identification problem with known variance. 

The algorithm has a simple and intuitive structure. However, we'd like to make a few important remarks. The initial phase goes in a round-robin manner. We deliberately avoid calling it an \textit{exploration} phase. The reason is that, if we think about the \textit{exploration} in multi-armed bandits literature, it generally refers to certain actions taken to collect data/information in sacrifice of short-term reward. Specifically, any play of suboptimal arms in multi-armed bandits will incur regret, but such plays are inevitable if we want to learn the system. Yet, for the context of multi-group mean estimation, Proposition \ref{prop:optimal} says that the optimal allocation requires going with each alternative $\Omega(T)$ times. Thus, the initial phase of Algorithm \ref{alg:FETE} is not only to construct variance estimates, but these rounds of selections are indeed necessarily required by the optimal solution. That's why we call our algorithm exploration-free. The $\Omega(T)$ times of selections prescribed by the optimal solution give a sufficiently good estimation of the system, and no additional exploration is needed. In this light, our result tells that the UCB design in \citep{aznag2023an} is redundant. In addition, we want to compare multi-group mean estimation with the problem of best arm identification. For both problems, they have an objective function different from the regret in multi-armed bandits. A special point is that the objective function $R_p(\bm{n})$ of multi-group mean estimation is closed-form in terms of the allocation scheme $\bm{n}$, whereas the probability of correct selection has a complicated relation with the allocation scheme (except for simple cases like two-armed bandits). The closed-formedness is the key to admitting simple algorithms like Algorithm \ref{alg:FETE}.

\subsection{Analysis of Algorithm \ref{alg:FETE}}

To facilitate our presentation, we define
\begin{align*}
\Sigma_{q} \coloneqq \sum_{k=1}^{K} \sigma_{k}^{q}, \quad
\lambda \coloneqq  \frac{\underline{\sigma}^q}{\underline{\sigma}^q+(K-1)\cdot \sigma^q} = \frac{\tau}{T},
\end{align*}
where $q$ is determined by $p$ as in Proposition \ref{prop:optimal}. We denote the allocation scheme of Algorithm \ref{alg:FETE} as $\bm{n}_{\pi_1}$ that represents the number of times each alternative is selected by the end of the horizon under Algorithm \ref{alg:FETE}. Let $\boldsymbol{\sigma^{2}} \coloneqq \{\sigma^{2}_{1}, \cdots, \sigma^{2}_{K}\}$ for simplicity. The following theorems give the bounds for the case of $p=\infty$ and $p<\infty$ respectively.

\begin{theorem} \label{thm:pinf}
For $p=\infty$, we have 
\begin{align*}
    \mathbb{E}\left[R_{p}(\bm{n}_{\pi_1}) - R_{p}(\bm{n}^{*})\right] &\leq 4\sqrt{2}\sigma^{2}\FalgInf{1}{\lambda,\boldsymbol{\sigma^{2}}}T^{-3/2}\sqrt{\log T}+o(T^{-3/2}),
    \end{align*}
    where $\FalgInf{1}{\lambda,\boldsymbol{\sigma^{2}}} \coloneqq \lambda^{-1/2}(K+\Sigma_{2}/\underline{\sigma}^{2}-2)$.
\end{theorem}

\begin{theorem} \label{thm:pfinite}
For $p<\infty$ ,we have:
    \begin{align*}
        \mathbb{E}\left[R_{p}(\bm{n}_{\pi_1}) - R_{p}(\bm{n}^{*})\right] &\leq 24\sigma^{4}\Falgp{1}{\lambda, \boldsymbol{\sigma^{2}}}T^{-2}\log T+o(T^{-2}),
    \end{align*}
    where $\Falgp{1}{\lambda, \boldsymbol{\sigma^{2}}} \coloneqq \frac{p^{2}(\Sigma_{q})^{1/p}\Sigma_{q-4}}{\lambda(p+1)}$.
\end{theorem}

The analyses for the case of infinite and finite $p$ are largely similar, with minor differences caused by the optimality structure of $\bm{n}^*.$ For both cases, we note that the optimal value $R_p(\bm{n}^*)$ is on the order of $1/T;$ therefore, the algorithm is asymptotically optimal given the optimality gap on the order of $T^{-3/2}$ and $T^{-2}.$ We make the following remarks about the results. First, the bounds improve on orders of $\log T$ compared to the respective results in \citep{aznag2023an}. The main differences of our algorithm and analysis are (i) the refined variance concentration inequality in Lemma \ref{lemma:variance_concentrarion} and (ii) getting rid of the UCB design. This reinforces our point that solving the problem of multi-group mean estimation can be exploration-free. In addition, we emphasize that the bounds can be further refined under the case of strictly subgaussian or gaussian, which we defer to Appendix~\ref{appendix:strictlysubgaussian}. 

\section{General Case}

In the previous section, we consider the case where a lower bound $\underline{\sigma}^2$ for the variances is known a priori. Now we consider the general case where there is no such prior knowledge. We note that in Algorithm \ref{alg:FETE} the only point where we use $\underline{\sigma}^2$ is to determine the length of the initial phase $\tau$. The knowledge of $\underline{\sigma}^2$ ensures that we will not exhaust the optimal budget $n_k^*$ in the initial phase. Thus, the idea of our second algorithm is to replace the knowledge of $\underline{\sigma}^2$ with some variance estimate based on the collected observations. Accordingly, the allocation scheme will be more adaptively adjusted based on the data flow. To simplify the notations, suppose some LCB and UCB estimates for the variances satisfy 
$$\mathbb{P}\left(\text{LCB}_{k,n}\le \sigma_k^2 \le \text{UCB}_{k,n}\text{ for all }k\text{ and }n \right) \ge 1 - 2T^{-c},$$
where $c$ will be determined by $p$. Here the event is taken as a union over all the alternatives $k$ and all the number of observations $n$ (up to $T$). Specifically, for each alternative $k$, one can construct such confidence bounds of \text{LCB}$_{k,n}$ and \text{UCB}$_{k,n}$ based on the sample variance estimator and Lemma \ref{lemma:variance_concentrarion}. The width of the confidence interval can be adjusted to ensure that the \textit{good} event (of true variances falling in confidence bounds uniformly) happens with a high probability. The detailed components of the LCBs and UCBs and the value of $c$ are deferred to Appendix~\ref{appendix:section4}.

\begin{algorithm}[ht!]
\caption{Adaptive Algorithm}
\begin{algorithmic}[1]
\State \textbf{Input:} Time horizon $ T $,  constant $q$, constant $m$.
\State \textbf{Phase 1: Avoid a meaningless LCB}
\State Select each alternative $n=O(1)$ times such that
$$\min_{k=1,...,K} \text{LCB}_{k,n}>0$$
\State \textbf{Phase 2: Determine stopping times}
\State Initialize \( \mathcal{A}_{\text{active}} \leftarrow \{1, \dots, K\} \)
\State Initialize \( n_{k} \leftarrow n \), \( \lambda_{k,n_{k}} \leftarrow n/T \).
\Repeat
    \For{each \( k \in \mathcal{A}_{\text{active}} \)}
        \State Select \( k \) for $\lambda_{k,n_{k}}T - n_{k}$ times
        \State Update \( n_{k} \leftarrow \lambda_{k,n_{k}} T\)
    \EndFor
    \State Compute \( \text{LCB}_{k,n_{k}} \), \( \text{UCB}_{k,n_{k}} \) for all $k$.
    \For{each \( k = 1, \dots, K \)}
        \State Update: $$\lambda_{k,n_{k}} = \frac{\text{LCB}^{q/2}_{k,n_{k}}}{\text{LCB}^{q/2}_{k,n_{k}}+\sum_{j\neq k}\text{UCB}^{q/2}_{j,n_{j}}}$$  
        \If{ \( n_{k} \geq \lambda_{k,n_{k}}  T \) }
            \State Set \( \tau_{k} \leftarrow n_{k} \), and remove  \( k \) from \( \mathcal{A}_{\text{active}} \)
        \Else
            \State Add \( k \) back to \( \mathcal{A}_{\text{active}} \) if $k \notin \mathcal{A}_{\text{active}}$
        \EndIf
    \EndFor
\Until{ \( \mathcal{A}_{\text{active}} = \emptyset \) } 

\State \textbf{Phase 3: Allocate the remaining periods}
\For{each  \( k = 1, \cdots, K \)}
    \State Compute \( \hat{\sigma}^{2}_{k,\tau_{k}} \) and calculate \( \lambda_{k,\tau_{k}} \) with Equation~\eqref{eq:allocation_lambda}
    \State Select \( k \) for \( \lambda_{k,\tau_{k}} T - \tau_{k} \) rounds
    \State Calculate: $$\hat{\mu}_{k}(T) = \frac{1}{\lambda_{k,\tau_{k}}T}\sum_{s=1}^{\lambda_{k,\tau_{k}}T} X_{k,s}$$
\EndFor

\State \textbf{Output:} Final estimates \( \{\hat{\mu}_{k}(T)\}_{k=1}^{K}\).
\end{algorithmic}
\label{alg:adaptiveETC}
\end{algorithm}

Algorithm~\ref{alg:adaptiveETC} presents our adaptive algorithm for the general case, i.e., without the knowledge of $\underline{\sigma}^2$. Phase 1 of the algorithm is a trivial part that simply aims to ensure all the LCB estimates are positive. The key of Phase 2 is the quantity $\lambda_{k,n_k}$, which implies a lower bound for $n_{k}^*$. The active set $\mathcal{A}_{\text{active}}$ maintains the alternatives that still need some more rounds of selection towards the optimal allocation. The quantity $\lambda_{k,n_k}$ is closely related the constant $\lambda$ and the initial length $\tau$ in Algorithm \ref{alg:FETE} where we replace LCBs and UCBs with $\underline{\sigma}^2$ and $\sigma^2$. While both designs aim to ensure that we don't over-select an alternative, LCBs and UCBs are more adaptive to the data, and thus Algorithm \ref{alg:adaptiveETC} should give a better performance. The last phase of the algorithm simply exhausts the remaining time steps as the second phase of Algorithm \ref{alg:FETE}.

In the literature of multi-armed bandits and best-arm identification, there is a line of works \citep{auer2010ucb, karnin2013almost, soare2014best, qian2016randomized} that utilize adaptive arm elimination (as Phase 2 of Algorithm \ref{alg:adaptiveETC}). Among this stream of works and algorithms, our arm elimination procedure is most similar to Procedure~1 in \citet{cai2024transfer} which focuses on the reward objective, and the algorithm in \citet{li2021symmetry} which also incorporates LCBs/UCBs into the optimization problem to perform arm elimination. 

\begin{theorem}\label{thm:regretAlg3_subgaussian}
We have the following results for Algorithm~\ref{alg:adaptiveETC}. For $p=\infty$, 
    \begin{align*}
        \mathbb{E}\left[R_{p}(\bm{n}_{\pi_2}) - R_{p}(\bm{n}^{*})\right] &\leq 8\sigma^{2}\FalgInf{2}{\boldsymbol{\sigma^{2}}}T^{-3/2}\sqrt{\log T}+o(T^{-3/2}),  
    \end{align*}
and for $p$ is finite, we have:
    \begin{align*}
        \mathbb{E}\left[R_{p}(\bm{n}_{\pi_2}) - R_{p}(\bm{n}^{*})\right] &\leq 40\sigma^{4} \Falgp{2}{\boldsymbol{\sigma^{2}}}T^{-2}\log T + o(T^{-2}),
    \end{align*}
where $\Falgp{2}{\boldsymbol{\sigma^{2}}} \coloneqq \frac{p^{2}(\Sigma_{q})^{2/q}(\Sigma_{-4})}{p+1}$, $\FalgInf{2}{\boldsymbol{\sigma^{2}}} \coloneqq \sqrt{\Sigma_{2}}\!\left(\Sigma_{-1}+\frac{\Sigma_{2}}{\sigma_{\min}^{3}}-\frac{2}{\sigma_{\min}}\right)$, and $\bm{n}_{\pi_2}$ denotes the allocation scheme of Algorithm \ref{alg:adaptiveETC}.
\end{theorem}

Theorem \ref{thm:regretAlg3_subgaussian} gives the performance bounds for Algorithm \ref{alg:adaptiveETC}. The analysis shares a similar spirit with that of Algorithm \ref{alg:FETE} but it deals with some additional complications caused by the LCBs and UCBs. We make several remarks for Algorithm \ref{alg:adaptiveETC} and Theorem \ref{thm:regretAlg3_subgaussian}. First, as in the case of Algorithm \ref{alg:FETE}, though Algorithm \ref{alg:adaptiveETC} involves the elements of LCBs and UCBs, it is still exploration-free. In other words, LCBs and UCBs arise from inaccurate estimates of the variance, but they don't incur any redundant play of any alternative $k$. Second, Algorithm \ref{alg:adaptiveETC} gives a bound on the same order as Algorithm \ref{alg:FETE} under the scenario of no prior knowledge on $\underline{\sigma}^2.$ Essentially, one can establish (from the analyses of these two algorithms) that as long as (i) each alternative is played $\Omega(T)$ (ii) the number of plays doesn't exceed the optimal scheme $n^*_k$, then one can always achieve an optimal gap as the ones in Theorems \ref{thm:pinf}, \ref{thm:pfinite}, and \ref{thm:regretAlg3_subgaussian}. Either the knowledge of $\underline{\sigma}^2$ or the LCB/UCB design in Algorithm \ref{alg:adaptiveETC} is used to ensure these two conditions. Lastly, we note that a strictly subgaussian distribution will give better bounds both theoretically and numerically. In particular, it eliminates the need for prior knowledge of $\sigma^{2}$ in constructing LCBs and UCBs, and further reduces the sampling requirement in the first phase of Algorithm~\ref{alg:adaptiveETC} to a small constant that depends only on $T$ and $c$. We defer to Appendix~\ref{appendix:strictlysubgaussian} for a detailed discussion.

\section{Contextual Case}

In this section, we extend the multi-group estimation to a contextual bandits setting where the goal is to estimate the group-level linear parameters rather than just the means of rewards. We adopt the multiple linear contextual bandit model described in~\citet{slivkins2019introduction, slivkins2011contextual} and follow the notations in the previous sections to ensure consistency throughout the paper. 

\subsection{Problem Setting}
Consider a contextual bandit setting with \( K \) arms, where each arm \( k \in \{1, \dots, K\} \) is associated with an unknown parameter vector \( \beta_k \in \mathbb{R}^d \). At each round \( t = 1, \dots, T \) (assuming $d \ll T$), a context vector \( c_t \in \mathbb{R}^d \) is observed, drawn i.i.d. from a distribution \( \mathcal{P}_\mathcal{C} \). In other words, we consider the setting of \textit{stochastic context}. For \( \mathcal{P}_\mathcal{C} \), we make the following assumptions.
\begin{assumption}\label{assump:postive_definite}
    We assume $\Sigma \coloneqq \mathbb{E}_{c\sim\mathcal{P}_{\mathcal{C}}}\left[c c^\top\right] \succ 0$. Moreover,
   $\|c_n\|\le R < \infty$ almost surely.
\end{assumption}

Let $\lambda_{\min}^{\mathcal{C}}\coloneqq\lambda_{\min}(\Sigma)$ represent the minimum and maximum eigenvalue respectively. Upon pulling arm \( k \), the observed reward is given by
\[
X_{k,n} = \beta_k^\top c_n + \eta_{k,n},
\]
where \( \eta_{k,n} \) is zero-mean noise, i.i.d. across time and arms, but with unknown arm-dependent variances \(\sigma^{2}_{k}\). And we assume that  $\eta_{k,n}$'s are subgaussian and satisfy Assumption~\ref{assump:subgaussian}. The goal is to determine the number of times needed for each arm beforehand to estimate the parameter vector \( \beta_k \) for each arm, and to evaluate the overall estimation error using the squared \( \ell_2 \)-norm:
\begin{align}
&\min \ \ \mathbb{E}_{ \!\ \mathcal{C},\eta}\left[\sum_{k=1}^K \|\hat{\beta}_{k,n_{k}} - \beta_k\|^{2}\right]  \label{eqn:context_mgme} \\
&\text{s.t.} \quad \sum_{k=1}^{K} n_{k} = T. \notag
\end{align}
where $\hat{\beta}_{k,n} = V^{-1}_{k,n}\sum_{s=1}^{n}c_{k,s}X_{k,s}^{\top}$, and $V_{k,n}=\gamma \cdot I_{d}+\sum_{s=1}^{n}c_{k,s}c_{k,s}^\top$ with $\gamma$ is the (ridge) penalty factor and $I_{d}$ is $d$-dimensional identity matrix. 
We remark that \citet{riquelme2017active} and \citet{fontaine2021online} consider a similar objective to ours. The optimization problem above can be viewed as a natural extension of that for the multi-group mean estimation.

\subsection{Algorithm and Analysis}

First, the following lemma characterizes the estimation error. In contrast to the setting in the previous sections, the error involves the number of observations $n$ in a much more complicated manner. Specifically, we note that the inverse sample covariance matrix appears in the expression. This prevents the usage of UCB-type algorithms (such as \cite{aznag2023an}) for this contextual multi-group estimation; this is because a UCB-based algorithm can provide an error bound in the data-dependent norm $\|\cdot\|_{V_{k,n}^{-1}}$ but not in the Euclidean norm $\|\cdot\|$. The data-dependent norm suffices for deriving a standard regret bound with the help of the elliptical potential lemma but cannot be transformed to a bound for \eqref{eqn:context_mgme}.

\begin{lemma}\label{lemma:condition_context_MSE}
Let $\boldsymbol{C_{k,n}}=[c_{k,1},\cdots,c_{k,n}]$ be the first $n$ context vectors shown for arm $k$, then we have:
    \begin{align*}
        \mathbb{E}\left[\|\hat{\beta}_{k,n} - \beta_{k}\|^{2} \mid \boldsymbol{C_{k,n}} \right] = \sigma_{k}^{2}\operatorname{Tr}(V_{k,n}^{-1})+   \gamma^{2}\beta_{k}^{\top}V_{k,n}^{-2}\beta_{k}-\gamma\sigma_{k}^{2}\operatorname{Tr}(V_{k,n}^{-2}),
\end{align*}
where the expectation is taken w.r.t. the noises $\eta_{k,n}$'s.
\end{lemma}

In our algorithm problem, we use the ridge regression to estimate $\beta_k$'s as the design in the linear bandits literature. This prevents the singularity of the sample covariance matrix. Before we proceed, we first present the matrix concentration inequality. 

\begin{theorem}(Theorem~6.1.1 in \citet{tropp2015introduction})\label{thm:matrix_bern}
Let $\{ \boldsymbol{X}_i\}_{i=1}^n$ be independent, centered, self–adjoint random
matrices in $\mathbb{R}^{d\times d}$ with $\mathbb{E}[\boldsymbol{X}_i]= \boldsymbol{0}$ and $\|\boldsymbol{X}_i\| \le R$ a.s., let $\nu = \bigl\| \sum_{i=1}^n \mathbb{E}[\boldsymbol{X}_i^{\,2}]\bigr\|$, then for every $t\ge0$:
\begin{align*}
    \mathbb{P} \left\{\left\|\sum_{i=1}^n \boldsymbol{X}_i\right\| \ge t \right\} &\le 2d \cdot \exp\left(-\frac{t^2/2}{\nu + Rt/3}\right).
\end{align*}
\end{theorem}

As an implication, we can derive the following bound for the ridge regression estimator.

\begin{lemma}\label{lemma:MSE_bound}
Let $\gamma =\lambda_{\min}^{\mathcal{C}}/n$, and for $n\geq 2$, we have:
\begin{align*}
   \mathbb{E}\left[\|\hat{\beta}_{k,n} - \beta_k\|^{2}\right]
   \;\le\;
   \frac{2d\sigma_{k}^{2}}{n\lambda_{\min}^{\mathcal{C}}} + o(n^{-2}).
\end{align*}
\end{lemma}

From the lemma, we can approximately represent the objective (\ref{eqn:context_mgme}) with the following optimization problem. The rationale is that the difference between the following problem and (\ref{eqn:context_mgme}) is of a lower order. We point out that this approximation requires an additional condition for the algorithm design, which we will address after describing the algorithm.
\begin{align*}
    \min\quad & R_T(\bm{n}) =  \frac{2d}{\lambda_{\min}^{\mathcal{C}}}\sum_{k=1}^{K}\frac{\sigma_k^{2}}{n_k} \\
    \text{s.t.}\quad & \sum_{k=1}^{K}n_{k} = T
\end{align*}
where the decision variables are the allocation scheme $\bm{n}=(n_1,...,n_K).$ The optimal solution of this problem is $n^{*}_{k} = \frac{\sigma_{k}}{\Sigma_{1}} \cdot T$. Now the problem reduces to estimating $\sigma^{2}_{k}$, which is similar to the group-mean estimation setup in the previous sections. Then we can define the residual term as:
\begin{align*}
    r_{k,s} = X_{k,s} - \hat{\beta}^{\top}_{k,n}c_{k,s}
\end{align*}
and the estimated variance becomes:
\begin{align} \label{eq:contextual_variance}
    \hat{\sigma}^{2}_{k,n} = \frac{1}{n-1}\sum_{s=1}^{n}(r_{k,s}-\frac{1}{n}\sum_{s=1}^{n}r_{k,s})^{2}
\end{align}

\begin{algorithm}[H]
\caption{Contextual Algorithm}
\begin{algorithmic}[1]
\State \textbf{Input:} Time horizon \( T \), context distribution \(\mathcal{P}_{\mathcal{C}}\), context dimension $d$, minimum eigenvalue $\lambda_{\min}^{\mathcal{C}}$.
\State Play each arm $d$ times $\{X_{k,n}\}_{n=1}^{d}$ and $\{c_{k,n}\}_{n=1}^{d}$
\State \textbf{Phase 1: Avoid a meaningless LCB}
\State Select each alternative $n=O(1)$ times such that
$$\min_{k=1,...,K} \text{LCB}_{k,n}>0$$
\State \textbf{Phase 2: Adaptive Elimination Strategy}
\State Employ the second phase of Algorithm~\ref{alg:adaptiveETC} until{ \( \mathcal{A}_{\text{active}} = \emptyset \) } based on $\hat{\sigma}^{2}_{k,n}$ with Equation~\eqref{eq:contextual_variance}. 
\State \textbf{Phase 3: Allocate the remaining periods}
\For{each arm \( k = 1, \cdots, K \)}
    \State Compute \( \hat{\sigma}^{2}_{k,n_{k}} \) and \( \lambda_{k,n_{k}} \)
    \State Play arm \( k \) for \( \lambda_{k,n_{k}} \cdot T - n_{k} \) rounds
    \State Calculate: $$\hat{\beta}_{k}(T) = V^{-1}_{k,n_{k}}\sum_{s=1}^{n_{k}}c_{k,s}X_{k,s}^{\top}$$
\EndFor
\State \textbf{Output:} Final estimates \( \{\hat{\beta}_{k}(T)\}_{k=1}^{K} \)
\end{algorithmic}
\label{alg:contextualETC}
\end{algorithm}

The algorithm follows an almost identical structure with Algorithm \ref{alg:adaptiveETC} by replacing the variance estimates by \eqref{eq:contextual_variance}. An important design of the algorithm is that in Phase 2, the allocation decision (which arm to select at time $t$) is decided before seeing the context $c_t$. This design is quite different from other algorithms on linear bandits. On one hand, this is admitted by the nature of the group-mean estimation, which requires each arm to be played for $\Omega(T)$ times. On the other hand, this first-decide-then-observe structure ensures the independence between context vectors shown for each arm, and hence makes Theorem \ref{thm:matrix_bern} and Lemma \ref{lemma:MSE_bound} applicable. The following theorem gives the performance bound for Algorithm \ref{alg:contextualETC}, which is comparable to the finite-$p$ case in the previous sections.

\begin{theorem}\label{thm:Contextual_regret}
    For Algorithm \ref{alg:contextualETC}, with $p=1$, we have: 
    \begin{align*}
        \mathbb{E}[R_T(\bm{n}_{\pi_{3}}) - R_T(\bm{n}^{*})] &\leq \frac{80d\sigma^{2}}{\lambda^{\mathcal{C}}_{\min}}\Falgp{3}{\boldsymbol{\sigma^{2}}}T^{-2}\log T + o(T^{-2}),
    \end{align*}
    where $\bm{n}_{\pi_3}$ denotes the allocation scheme of Algorithm~\ref{alg:contextualETC}, and $\Falgp{3}{\boldsymbol{\sigma^{2}}} = \Falgp{2}{\boldsymbol{\sigma^{2}}}$.
\end{theorem}

\section{Numerical Experiments}

In this section, we present numerical experiments for our algorithms where all the results are reported based on 100 simulation trials. We also refer to Appendix~\ref{appendix:experiments} for more experiments and details.

\subsection{Gaussian alternatives}
\begin{figure}[htbp]
  \centering
  \begin{subfigure}[t]{0.45\columnwidth}
      \centering
      \includegraphics[width=\linewidth]{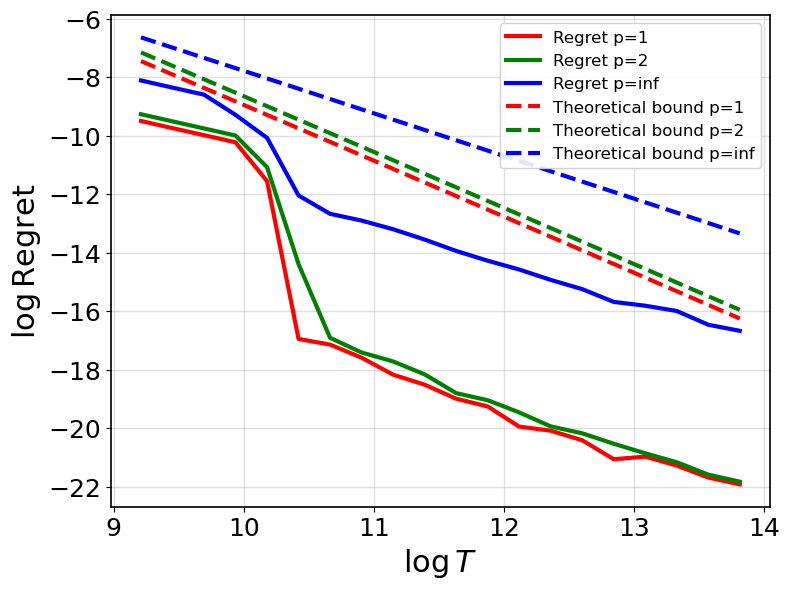}
      \caption{GSG without $\underline{\sigma}^2$.}
      \label{fig:sg_unknown}
  \end{subfigure}\hfill
  \begin{subfigure}[t]{0.45\columnwidth}
      \centering
      \includegraphics[width=\linewidth]{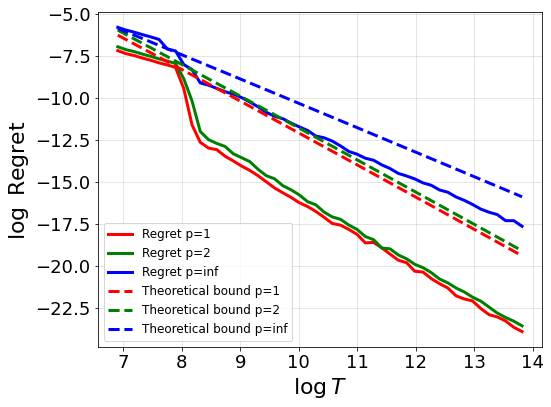}
      \caption{SSG without $\underline{\sigma}^2$.}
      \label{fig:ssg_unknown}
  \end{subfigure}\hfill
  \caption{Algorithm~\ref{alg:adaptiveETC} under two settings.}
  \label{fig:adaptive_Gaussian}
\end{figure}

For traditional bandit problems, we play $K=4$ arms generated from Gaussian distribution $\mathcal{G}_{k}$ with mean $\mu_{k} \sim \mathcal{U}([-1,1])$ and $\{\sigma^{2}_{1}, \sigma^{2}_{2}, \sigma^{2}_{3}, \sigma^{2}_{4} \} = \{1, 1.5, 2,2.5\}$ respectively. In this setting, we have $\sigma^2=2.5$ and $\underline{\sigma}^{2} = 1$ if known. We conduct the experiment under the general subgaussian (GSG) setting and the strictly subgaussian (SSG) setting, respectively. From Figure~\ref{fig:adaptive_Gaussian}, we observe a sharp performance drop when $\underline{\sigma}^{2}$ is unknown. Moreover, the point at which this drop occurs differs between the GSG and SSG settings. 
This discrepancy is caused by different lengths of Phase 1 in Algorithm~\ref{alg:adaptiveETC}. In the GSG setting, this length necessitates a large time horizon $T$, more than $2\times 10^{4}$. When $T$ is not large enough, the effective exploration budget per arm is only about $T/K$. Besides, we note that the theoretical upper bound in the SSG setting is much closer to the empirical regret, particularly when $p=\infty$, which corroborates the sharper guarantees predicted by our analysis. More visualizations are deferred to  Appendix~\ref{appendix:experiments_Gaussian}.

\subsection{Rademacher and Gaussian alternatives}

\begin{figure}[ht!]
    \centering
    \includegraphics[scale=0.4]{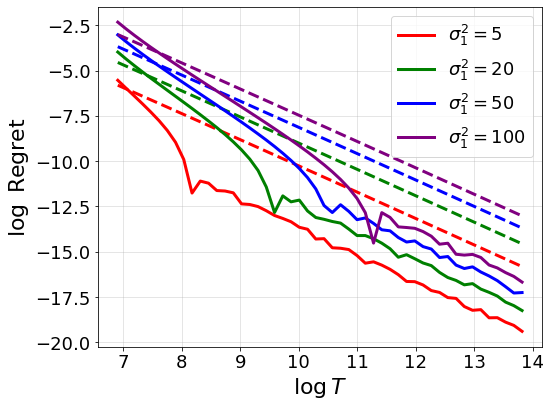}
\caption{Algorithm~\ref{alg:adaptiveETC}: Rademacher and Gaussian alternatives in SSG setting.}
\label{fig:rademacher}
\end{figure}

Now we reproduce the numerical experiment of Non--Gaussian arms in \citet{carpentier2011upper}. This is a two--arm bandit problem: one with a Gaussian $\mathcal{N}(0,\sigma_{1}^{2})$ with $\sigma_{1}^{2} \geq 1$, and another with Rademacher arm. They used $\sigma^{2} = \sigma^{2}_{1}+1$ as the prior information and $\lambda_{\min} = 1/(1+\sigma^{2}_{1})$.
In their experiment, they showed the $p$--norm result with $p=\infty$ of $T=10^{3}$ for different $\sigma_{1}^{2}$. In the GSG setting, the length of the first phase of Algorithm~\ref{alg:adaptiveETC} is related to $\sigma^{2}$. For a large $\sigma_{1}^{2}$ a sufficiently large horizon $T$ is required to realize the theoretical guarantees. By contrast, in the SSG setting the first-phase length reduces to a constant independent of $\sigma$, leading to a faster and simpler procedure. Since both the Gaussian and Rademacher distributions belong to the class of strictly subgaussian distributions, we conduct experiments in the SSG setting with $\sigma_{1}^{2} \in \{5, 20, 50, 100\}$. Figure~\ref{fig:rademacher} shows the empirical regret (solid line) compared against the theoretical upper bound (dashed line) for different $\sigma^{2}_{1}$. 

\subsection{Contextual Bandit}
\begin{figure}[htbp]
  \centering
  \begin{subfigure}[t]{0.45\columnwidth}
    \centering
    \includegraphics[width=\linewidth]{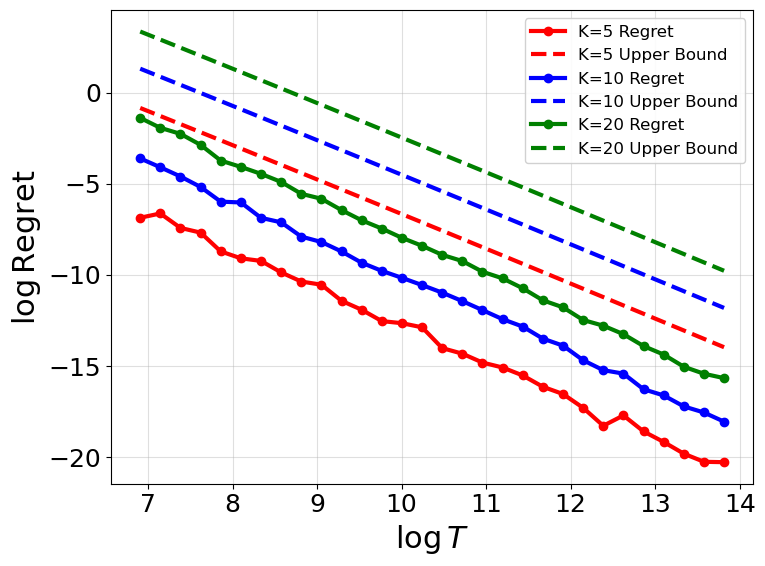}
    \caption{GSG setting}
    \label{fig:contextual_K_GSG}
  \end{subfigure}\hfill
  \begin{subfigure}[t]{0.45\columnwidth}
    \centering
    \includegraphics[width=\linewidth]
    {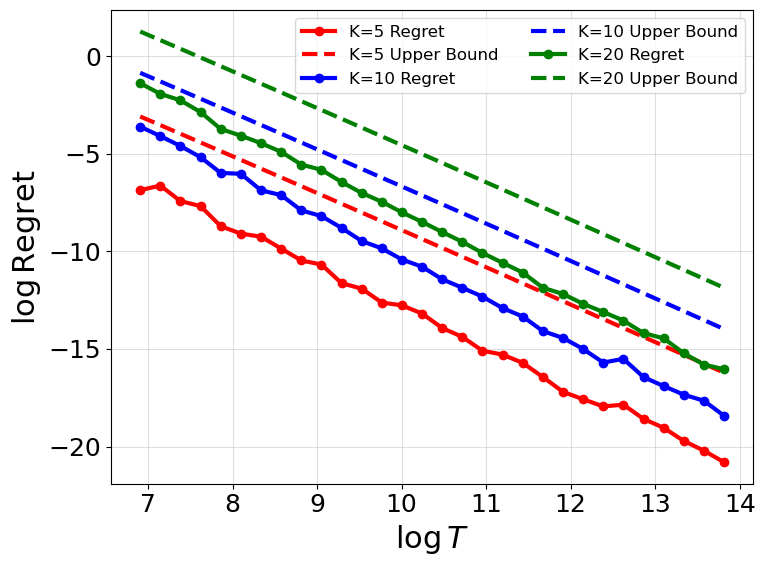}
    \caption{SSG setting}
    \label{fig:contextual_K_SSG}
  \end{subfigure}\hfill
  \caption{Algorithm~\ref{alg:contextualETC}: Contextual setting.}
  \label{fig:contextual_K}
\end{figure}

We consider a contextual bandit setting with varying numbers of arms $K \in \{5,10,20\}$ and dimension $d=4$. Context vectors are sampled uniformly from the hypercube $\left[-\sqrt{3},\,\sqrt{3}\right]^{d}$. 
For each arm $k$, the coefficient vector $\beta_{k}$ is independently drawn from $\mathcal{U}[-2,2]^{d}$, 
while the noise $\eta_{k}$ follows a Gaussian distribution with variance sampled from $\mathcal{U}[1,4]$. We assume $\underline{\sigma}^{2}=1$ and $\sigma^{2}=4$, and evaluate the performance under both the GSG and SSG settings.  As shown in Figure~\ref{fig:contextual_K}, the empirical regret exhibits a slope close to $-2$, which is consistent with the rate predicted by Theorem~\ref{thm:Contextual_regret}.

\section{Conclusion and Discussions}
In this paper, we study the multi-group mean estimation problem and consider both the canonical and the contextual settings of the problem. We propose several algorithm that features a simpler design than the existing ones but achieves the optimal order of regret. The proposed algorithms and analyses reveal several key structural insights to the problem: First, the optimal allocation scheme requires $\Omega(T)$ selections of each alternative/arm, and thus it enables exploration-free algorithms. Second, we point out a connection between the problem and the best arm identification problem, where the optimal allocation scheme is not in closed form. Third, we explain why the UCB-type exploration is unnecessary for the canonical setting and how it completely fails in the contextual setting. As a side product, we find that a strictly subgaussian distribution allows a sharper theoretical bound and also a better numerical performance. It suggests a potential research direction: If the reward distributions \( \mathcal{P}_{k} \) belong to such structured families, but the variances are only partially available, one could develop adaptive algorithms that exploit this structure. Such algorithms could iteratively refine their variance estimates and thereby achieve tighter confidence intervals, leading to further improved sample efficiency.


\bibliographystyle{informs2014}
\bibliography{main}

\begin{thebibliography}{54}
\expandafter\ifx\csname natexlab\endcsname\relax\def\natexlab#1{#1}\fi
\expandafter\ifx\csname url\endcsname\relax
  \def\url#1{{\tt #1}}\fi
\expandafter\ifx\csname urlprefix\endcsname\relax\def\urlprefix{URL }\fi
\expandafter\ifx\csname urlstyle\endcsname\relax
  \expandafter\ifx\csname doi\endcsname\relax
  \def\doi#1{doi:\discretionary{}{}{}#1}\fi \else
  \expandafter\ifx\csname doi\endcsname\relax
  \def\doi{doi:\discretionary{}{}{}\begingroup \urlstyle{rm}\Url}\fi \fi

\bibitem[{Agarwal et~al.(2009)Agarwal, Chen, Elango, Motgi, Park, Ramakrishnan, Roy, and Zachariah}]{agarwal2009online}
Agarwal, Deepak, Bee-Chung Chen, Pradheep Elango, Nitin Motgi, Seung-Taek Park, Raghu Ramakrishnan, Scott Roy, Joe Zachariah. 2009.
\newblock Online models for content optimization.
\newblock {\it Advances in Neural Information Processing Systems\/}. 17--24.

\bibitem[{Allen-Zhu et~al.(2021)Allen-Zhu, Li, Singh, and Wang}]{allen2021near}
Allen-Zhu, Zeyuan, Yuanzhi Li, Aarti Singh, Yining Wang. 2021.
\newblock Near-optimal discrete optimization for experimental design: A regret minimization approach.
\newblock {\it Mathematical Programming\/} {\bf 186}(1) 439--478.

\bibitem[{Amani et~al.(2019)Amani, Alizadeh, and Thrampoulidis}]{amani2019linear}
Amani, Sanae, Mahnoosh Alizadeh, Christos Thrampoulidis. 2019.
\newblock Linear stochastic bandits under safety constraints.
\newblock {\it Advances in Neural Information Processing Systems\/} {\bf 32}.

\bibitem[{Antos et~al.(2008)Antos, Grover, and Szepesv{\'a}ri}]{antos2008active}
Antos, Andr{\'a}s, Varun Grover, Csaba Szepesv{\'a}ri. 2008.
\newblock Active learning in multi-armed bandits.
\newblock {\it Proceedings of the 19th International Conference on Algorithmic Learning Theory (ALT)\/}, {\it Lecture Notes in Computer Science\/}, vol. 5254. Springer, 287--302.

\bibitem[{Antos et~al.(2010)Antos, Grover, and Szepesv{\'a}ri}]{antos2010active}
Antos, Andr{\'a}s, Varun Grover, Csaba Szepesv{\'a}ri. 2010.
\newblock Active learning in heteroscedastic noise.
\newblock {\it Theoretical Computer Science\/} {\bf 411}(29-30) 2712--2728.

\bibitem[{Arbel et~al.(2020)Arbel, Marchal, and Nguyen}]{arbel2020strict}
Arbel, Julyan, Olivier Marchal, Hien~D Nguyen. 2020.
\newblock On strict sub-gaussianity, optimal proxy variance and symmetry for bounded random variables.
\newblock {\it ESAIM: Probability and Statistics\/} {\bf 24} 39--55.

\bibitem[{Audibert and Bubeck(2010)}]{audibert2010best}
Audibert, Jean-Yves, S{\'e}bastien Bubeck. 2010.
\newblock Best arm identification in multi-armed bandits.
\newblock {\it COLT-23th Conference on learning theory-2010\/}. 13--p.

\bibitem[{Auer et~al.(2002)Auer, Cesa-Bianchi, and Fischer}]{auer2002finite}
Auer, Peter, Nicol{\`o} Cesa-Bianchi, Paul Fischer. 2002.
\newblock Finite-time analysis of the multiarmed bandit problem.
\newblock {\it Machine Learning\/} {\bf 47}(2-3) 235--256.

\bibitem[{Auer and Ortner(2010)}]{auer2010ucb}
Auer, Peter, Ronald Ortner. 2010.
\newblock Ucb revisited: Improved regret bounds for the stochastic multi-armed bandit problem.
\newblock {\it Periodica Mathematica Hungarica\/} {\bf 61}(1-2) 55--65.

\bibitem[{Aznag et~al.(2023)Aznag, Cummings, and Elmachtoub}]{aznag2023an}
Aznag, Abdellah, Rachel Cummings, Adam~N Elmachtoub. 2023.
\newblock An active learning framework for multi-group mean estimation.
\newblock {\it Advances in Neural Information Processing Systems\/} {\bf 36} 32602--32635.

\bibitem[{Ban and Keskin(2021)}]{ban2021personalized}
Ban, Gah-Yi, N~Bora Keskin. 2021.
\newblock Personalized dynamic pricing with machine learning: High-dimensional features and heterogeneous elasticity.
\newblock {\it Management Science\/} {\bf 67}(9) 5549--5568.

\bibitem[{Bastani and Bayati(2020)}]{bastani2020online}
Bastani, Hamsa, Mohsen Bayati. 2020.
\newblock Online decision making with high-dimensional covariates.
\newblock {\it Operations Research\/} {\bf 68}(1) 276--294.

\bibitem[{Bastani et~al.(2021)Bastani, Bayati, and Khosravi}]{bastani2021mostly}
Bastani, Hamsa, Mohsen Bayati, Khashayar Khosravi. 2021.
\newblock Mostly exploration-free algorithms for contextual bandits.
\newblock {\it Management Science\/} {\bf 67}(3) 1329--1349.

\bibitem[{Bobkov et~al.(2024)Bobkov, Chistyakov, and G{\"o}tze}]{bobkov2024strictly}
Bobkov, SG, GP~Chistyakov, Friedrich G{\"o}tze. 2024.
\newblock Strictly subgaussian probability distributions.
\newblock {\it Electronic Journal of Probability\/} {\bf 29} 1--28.

\bibitem[{Boucheron et~al.(2013)Boucheron, Lugosi, and Massart}]{boucheron2013concentration}
Boucheron, S{\'e}bastien, G{\'a}bor Lugosi, Pascal Massart. 2013.
\newblock {\it Concentration Inequalities: A Nonasymptotic Theory of Independence\/}.
\newblock Oxford University Press.

\bibitem[{Cai et~al.(2024)Cai, Cai, and Li}]{cai2024transfer}
Cai, Changxiao, T~Tony Cai, Hongzhe Li. 2024.
\newblock Transfer learning for contextual multi-armed bandits.
\newblock {\it The Annals of Statistics\/} {\bf 52}(1) 207--232.

\bibitem[{Carpentier et~al.(2011)Carpentier, Lazaric, Ghavamzadeh, Munos, and Auer}]{carpentier2011upper}
Carpentier, Alexandra, Alessandro Lazaric, Mohammad Ghavamzadeh, R{\'e}mi Munos, Peter Auer. 2011.
\newblock Upper-confidence-bound algorithms for active learning in multi-armed bandits.
\newblock {\it International Conference on Algorithmic Learning Theory\/}. Springer, 189--203.

\bibitem[{Dimakopoulou et~al.(2019)Dimakopoulou, Zhou, Athey, and Imbens}]{dimakopoulou2019balanced}
Dimakopoulou, Maria, Zhengyuan Zhou, Susan Athey, Guido Imbens. 2019.
\newblock Balanced linear contextual bandits.
\newblock {\it Proceedings of the AAAI Conference on Artificial Intelligence\/}, vol.~33. 3445--3453.

\bibitem[{Du et~al.(2024)Du, Gao, and Chen}]{du2024contextual}
Du, Jianzhong, Siyang Gao, Chun-Hung Chen. 2024.
\newblock A contextual ranking and selection method for personalized medicine.
\newblock {\it Manufacturing \& Service Operations Management\/} {\bf 26}(1) 167--181.

\bibitem[{Epperly(2022)}]{epperly2022hanson}
Epperly, Ethan. 2022.
\newblock Note to self: Hanson-wright inequality.
\newblock \urlprefix\url{https://www.ethanepperly.com/index.php/2022/10/04/note-to-self-hanson-wright-inequality}.

\bibitem[{Fontaine et~al.(2021)Fontaine, Perrault, Valko, and Perchet}]{fontaine2021online}
Fontaine, Xavier, Pierre Perrault, Michal Valko, Vianney Perchet. 2021.
\newblock Online a-optimal design and active linear regression.
\newblock {\it International Conference on Machine Learning\/}. PMLR, 3374--3383.

\bibitem[{Garcelon et~al.(2020)Garcelon, Ghavamzadeh, Lazaric, and Pirotta}]{garcelon2020improved}
Garcelon, Evrard, Mohammad Ghavamzadeh, Alessandro Lazaric, Matteo Pirotta. 2020.
\newblock Improved algorithms for conservative exploration in bandits.
\newblock {\it Proceedings of the AAAI Conference on Artificial Intelligence\/}, vol.~34. 3962--3969.

\bibitem[{Glynn and Juneja(2004)}]{glynn2004large}
Glynn, Peter, Sandeep Juneja. 2004.
\newblock A large deviations perspective on ordinal optimization.
\newblock {\it Proceedings of the 2004 Winter Simulation Conference, 2004.\/}, vol.~1. IEEE.

\bibitem[{Hao et~al.(2020)Hao, Lattimore, and Szepesvari}]{hao2020adaptive}
Hao, Botao, Tor Lattimore, Csaba Szepesvari. 2020.
\newblock Adaptive exploration in linear contextual bandit.
\newblock {\it International Conference on Artificial Intelligence and Statistics\/}. PMLR, 3536--3545.

\bibitem[{Karnin et~al.(2013)Karnin, Koren, and Somekh}]{karnin2013almost}
Karnin, Zohar, Tomer Koren, Oren Somekh. 2013.
\newblock Almost optimal exploration in multi-armed bandits.
\newblock {\it International conference on machine learning\/}. PMLR, 1238--1246.

\bibitem[{Kazerouni et~al.(2017)Kazerouni, Ghavamzadeh, Abbasi~Yadkori, and Van~Roy}]{kazerouni2017conservative}
Kazerouni, Abbas, Mohammad Ghavamzadeh, Yasin Abbasi~Yadkori, Benjamin Van~Roy. 2017.
\newblock Conservative contextual linear bandits.
\newblock {\it Advances in neural information processing systems\/} {\bf 30}.

\bibitem[{Khamaru et~al.(2021)Khamaru, Deshpande, Lattimore, Mackey, and Wainwright}]{khamaru2021near}
Khamaru, Koulik, Yash Deshpande, Tor Lattimore, Lester Mackey, Martin~J Wainwright. 2021.
\newblock Near-optimal inference in adaptive linear regression.
\newblock {\it arXiv preprint arXiv:2107.02266\/} .

\bibitem[{Lai and Robbins(1985)}]{lai1985asymptotically}
Lai, Tze~Leung, Herbert Robbins. 1985.
\newblock Asymptotically efficient adaptive allocation rules.
\newblock {\it Advances in applied mathematics\/} {\bf 6}(1) 4--22.

\bibitem[{Lattimore and Szepesv{\'a}ri(2020)}]{lattimore2020bandit}
Lattimore, Tor, Csaba Szepesv{\'a}ri. 2020.
\newblock {\it Bandit Algorithms\/}.
\newblock Cambridge University Press.

\bibitem[{Laurent and Massart(2000)}]{laurent2000adaptive}
Laurent, Béatrice, Pascal Massart. 2000.
\newblock Adaptive estimation of a quadratic functional by model selection.
\newblock {\it The Annals of Statistics\/} {\bf 28}(5) 1302--1338.

\bibitem[{Li et~al.(2010)Li, Chu, Langford, and Schapire}]{li2010contextual}
Li, Lihong, Wei Chu, John Langford, Robert~E Schapire. 2010.
\newblock A contextual-bandit approach to personalized news article recommendation.
\newblock {\it Proceedings of the 19th international conference on World wide web\/}. 661--670.

\bibitem[{Li et~al.(2021)Li, Sun, and Ye}]{li2021symmetry}
Li, Xiaocheng, Chunlin Sun, Yinyu Ye. 2021.
\newblock The symmetry between arms and knapsacks: A primal-dual approach for bandits with knapsacks.
\newblock {\it International Conference on Machine Learning\/}. PMLR, 6483--6492.

\bibitem[{Martinez-Taboada and Ramdas(2025)}]{martinez2025sharp}
Martinez-Taboada, Diego, Aaditya Ramdas. 2025.
\newblock Sharp empirical bernstein bounds for the variance of bounded random variables.
\newblock {\it arXiv preprint arXiv:2505.01987\/} .

\bibitem[{Maurer and Pontil(2009)}]{maurer2009empirical}
Maurer, Andreas, Massimiliano Pontil. 2009.
\newblock Empirical bernstein bounds and sample-variance penalization.
\newblock {\it Proceedings of the 22nd Annual Conference on Learning Theory (COLT)\/}. 115--124.

\bibitem[{Pukelsheim(2006)}]{pukelsheim2006optimal}
Pukelsheim, Friedrich. 2006.
\newblock {\it Optimal Design of Experiments\/}, {\it Classics in Applied Mathematics\/}, vol.~50.
\newblock SIAM.

\bibitem[{Qian and Yang(2016)}]{qian2016randomized}
Qian, Wei, Yuhong Yang. 2016.
\newblock Randomized allocation with arm elimination in a bandit problem with covariates.
\newblock {\it Electronic Journal of Statistics\/} {\bf 10}(1) 242--270.

\bibitem[{Qiang and Bayati(2016)}]{qiang2016dynamic}
Qiang, Sheng, Mohsen Bayati. 2016.
\newblock Dynamic pricing with demand covariates.
\newblock {\it arXiv preprint arXiv:1604.07463\/} .

\bibitem[{Rigollet and H{\"u}tter(2023)}]{rigollet2023high}
Rigollet, Philippe, Jan-Christian H{\"u}tter. 2023.
\newblock High-dimensional statistics.
\newblock {\it arXiv preprint arXiv:2310.19244\/} .

\bibitem[{Riquelme et~al.(2017)Riquelme, Ghavamzadeh, and Lazaric}]{riquelme2017active}
Riquelme, Carlos, Mohammad Ghavamzadeh, Alessandro Lazaric. 2017.
\newblock Active learning for accurate estimation of linear models.
\newblock {\it International Conference on Machine Learning\/}. PMLR, 2931--2939.

\bibitem[{Rudelson and Vershynin(2013)}]{rudelson2013hanson}
Rudelson, Mark, Roman Vershynin. 2013.
\newblock Hanson-wright inequality and sub-gaussian concentration.
\newblock {\it Electronic Communications in Probability\/} {\bf 18} 1--9.

\bibitem[{Sabato and Munos(2014)}]{sabato2014active}
Sabato, Sivan, Remi Munos. 2014.
\newblock Active regression by stratification.
\newblock {\it Advances in Neural Information Processing Systems\/} {\bf 27}.

\bibitem[{Shekhar et~al.(2020)Shekhar, Javidi, and Ghavamzadeh}]{shekhar20a}
Shekhar, Shubhanshu, Tara Javidi, Mohammad Ghavamzadeh. 2020.
\newblock Adaptive sampling for estimating probability distributions.
\newblock {\it International Conference on Machine Learning\/}. PMLR, 8687--8696.

\bibitem[{Shin et~al.(2019{\natexlab{a}})Shin, Ramdas, and Rinaldo}]{shin2019sample}
Shin, Jaehyeok, Aaditya Ramdas, Alessandro Rinaldo. 2019{\natexlab{a}}.
\newblock Are sample means in multi-armed bandits positively or negatively biased?
\newblock {\it Advances in Neural Information Processing Systems\/} {\bf 32}.

\bibitem[{Shin et~al.(2019{\natexlab{b}})Shin, Ramdas, and Rinaldo}]{shin2019bias}
Shin, Jaehyeok, Aaditya Ramdas, Alessandro Rinaldo. 2019{\natexlab{b}}.
\newblock On the bias, risk and consistency of sample means in multi-armed bandits.
\newblock {\it arXiv preprint arXiv:1902.00746\/} .

\bibitem[{Simchi-Levi and Wang(2025)}]{simchi2025multi}
Simchi-Levi, David, Chonghuan Wang. 2025.
\newblock Multi-armed bandit experimental design: Online decision-making and adaptive inference.
\newblock {\it Management Science\/} {\bf 71}(6) 4828--4846.

\bibitem[{Slivkins(2011)}]{slivkins2011contextual}
Slivkins, Aleksandrs. 2011.
\newblock Contextual bandits with similarity information.
\newblock {\it Proceedings of the 24th annual Conference On Learning Theory\/}. JMLR Workshop and Conference Proceedings, 679--702.

\bibitem[{Slivkins(2019)}]{slivkins2019introduction}
Slivkins, Aleksandrs. 2019.
\newblock Introduction to multi-armed bandits.
\newblock {\it Foundations and Trends® in Machine Learning\/} {\bf 12}(1--2) 1--286.

\bibitem[{Soare et~al.(2014)Soare, Lazaric, and Munos}]{soare2014best}
Soare, Marta, Alessandro Lazaric, R{\'e}mi Munos. 2014.
\newblock Best-arm identification in linear bandits.
\newblock {\it Advances in neural information processing systems\/} {\bf 27}.

\bibitem[{Tropp(2015)}]{tropp2015introduction}
Tropp, Joel~A. 2015.
\newblock An introduction to matrix concentration inequalities.
\newblock {\it Foundations and Trends{\textregistered} in Machine Learning\/} {\bf 8}(1-2) 1--230.

\bibitem[{van~der Vaart(1998)}]{vandervaart1998asymptotic}
van~der Vaart, Aad~W. 1998.
\newblock {\it Asymptotic Statistics\/}.
\newblock Cambridge Series in Statistical and Probabilistic Mathematics, Cambridge University Press.

\bibitem[{Vershynin(2018)}]{vershynin2018hdp}
Vershynin, Roman. 2018.
\newblock {\it High-Dimensional Probability: An Introduction with Applications in Data Science\/}.
\newblock Cambridge University Press.

\bibitem[{Wainwright(2019)}]{wainwright2019high}
Wainwright, Martin~J. 2019.
\newblock {\it High-dimensional statistics: A non-asymptotic viewpoint\/}.
\newblock Cambridge University Press.

\bibitem[{Wan et~al.(2022)Wan, Kveton, and Song}]{wan2022safe}
Wan, Runzhe, Branislav Kveton, Rui Song. 2022.
\newblock Safe exploration for efficient policy evaluation and comparison.
\newblock {\it International Conference on Machine Learning\/}. PMLR, 22491--22511.

\bibitem[{Wu et~al.(2016)Wu, Shariff, Lattimore, and Szepesv{\'a}ri}]{wu2016conservative}
Wu, Yifan, Roshan Shariff, Tor Lattimore, Csaba Szepesv{\'a}ri. 2016.
\newblock Conservative bandits.
\newblock {\it International Conference on Machine Learning\/}. PMLR, 1254--1262.

\end{thebibliography}

\appendix
\section{Related Work}
\label{appendix: literature}
\subsection{Variance Concentration}
Variance concentration is fundamental for assessing the reliability of empirical estimates. Classical Bernstein inequalities yield variance-sensitive bounds that improve over Hoeffding-type results. \citet{maurer2009empirical} introduced empirical Bernstein bounds that replace the true variance with its empirical estimate under boundedness assumptions. \cite{carpentier2011upper} extended this to unbounded distributions via a truncation--based technique. Recently, \citet{martinez2025sharp}
provided sharper empirical Bernstein bounds for bounded case. Further refinements arise from the Hanson--Wright inequality, which provides sharp tail bounds for quadratic forms of subgaussian vectors and underpins concentration results for covariance and design matrices. \citet{rudelson2013hanson} gave a modern proof with unspecified constants. For the notion of subgamma, \citet{laurent2000adaptive} gave a proof for Gaussian distribution, and \citet{boucheron2013concentration} followed the proof framework and introduced the notion of subgamma, based on which \cite{epperly2022hanson} gave a proof of Hanson--Wright inequality with specific constants. Besides subgamma, another closely related and widely--used notion is subexponential distribution.
\citet{vershynin2018hdp}, \citet{wainwright2019high}, \citet{rigollet2023high} gave detailed introduction of subexponential distribution.

\subsection{Group Mean Estimation in MAB}
Group mean estimation is an important problem in the MAB framework, as many applications require uniformly accurate estimates of all arm means rather than simply identifying the best arm. \citet{shin2019sample, shin2019bias} analyzed the statistical properties of sample means under adaptive sampling. In this setting, the goal is to allocate sampling resources strategically to minimize estimation error under limited feedback. A major line of research builds on UCB--based algorithms, which is based on \citet{lai1985asymptotically} and \citet{auer2002finite}. \cite{antos2008active, antos2010active} were among the first to formally study active learning in MABs with heteroscedastic noise. Their work analyzed optimal allocation strategies for minimizing estimation error under bounded reward assumptions. \cite{carpentier2011upper} mentioned this topic is related to pure-exploration and provided new regret bounds by proposing UCB--based algorithms under unbounded case. Recently, \cite{aznag2023an} revisited this classical allocation problem using a $p$-norm objective to capture different notions of group-level estimation quality, and derived the theoretical lower bound. Besides estimating the means of distributions in squared error sense, \citet{shekhar20a} considered four general distance measures: $\ell^{2}_{2}$, $\ell_{1}$, $f$-divergence, and separation distance. For linear models, especially linear bandit and contextual bandit, the estimation becomes to estimate the linear coefficient $\beta$. \citet{riquelme2017active} studied this problem setting in the multi-linear regression with the assumption that both noise and context are generated from Gaussian distribution. When the mean of arms is a linear combination with an unknown parameter, then the problem becomes an optimal experimental design problem \citep{pukelsheim2006optimal, sabato2014active, dimakopoulou2019balanced, allen2021near, khamaru2021near, fontaine2021online, simchi2025multi}. Specifically, \citet{fontaine2021online} considered this univariate unknown parameter under heterogeneous subgaussian noise with unit-norm context vectors.

\subsection{Exploration-free Algorithms} \citet{lattimore2020bandit} gave an introduction of pure exploration. For exploration-free algorithms, one of the important techniques in the exploration period is arm elimination, which is widely used as a core method for best arm identification \citep{auer2010ucb, audibert2010best, karnin2013almost, soare2014best, qian2016randomized}. In the contextual bandit setting, \citet{bastani2021mostly} introduced the notion of natural exploration and showed that a warm-start greedy policy can be near-optimal with little explicit exploration. \citet{hao2020adaptive} developed an optimization-based adaptive exploration scheme that tracks the instance-optimal sampling allocation and achieved instance-dependent asymptotic optimality, with sub-logarithmic regret under rich context distributions, spurring a growing line of follow-up work over the recent years. \citet{wan2022safe} studied safe exploration for policy evaluation and comparison, formulating data collection as a constrained design problem and deriving exploration policies that ensure safety while improving the statistical efficiency of off-policy evaluation.

\subsection{Conservative Bandits} Conservative bandits study safety-constrained exploration that keeps performance close to a baseline while learning. Unlike exploration-free algorithms, conservative methods do not eliminate exploration, they restrict it to the minimum required to satisfy safety, so under strong baselines or tight constraints they may appear nearly greedy while still performing essential, safety-driven probing \citep{wu2016conservative, kazerouni2017conservative, amani2019linear, garcelon2020improved}.

\section{Subgaussian  Variance Concentration}\label{appendix: subgaussian_subgamma}

Accurate variance estimation plays a central role in the design and analysis of the algorithms developed in this paper, particularly in the context of active learning and adaptive allocation strategies. This section presents a collection of theoretical results that characterize the concentration behavior of empirical variance estimators, we review and extend several classical results under different distribution assumptions. 

\subsection{Preliminaries and Key Definitions}
In the context of variance estimation, it is important to note that both the sample mean and the sample variance can be formulated as U-statistics. A key theoretical property of U-statistics is that they satisfy a central limit theorem, which ensures their asymptotic normality under mild regularity conditions \citep{vandervaart1998asymptotic}. 

From Definition~\ref{def:subgaussian} of subgaussian, it is straightforward to verify that \(\mathbb{E}[X]=0\) and the variance \( \sigma_{X}^{2} \leq \sigma^2 \). A particularly interesting case arises when equality holds, i.e., \( \sigma^2 = \sigma_{X}^{2} \). This condition is satisfied by a special class of distributions named strictly--subgaussian distribution \citep{arbel2020strict, bobkov2024strictly}. It includes examples such as Gaussian, Rademacher, symmetric Beta distribution and so on.

While for some specific distribution types, symmetric distributions is crucial for strictly--subgaussian, but generally, symmetry is neither necessary nor sufficient for strictly--subgaussian, see Proposition~1.1 and 1.2 in \citet{arbel2020strict}. When $X$ is strictly--subgaussian with variance $\sigma^{2}_{X}$, we could get that $\mathbb{E}[X^{3}]=0$ and $\mathbb{E}[X^{4}]\leq 3\sigma_{X}^{4}$ by using Taylor expansion in Equation~\eqref{eq:subgaussian_definition}. We adopt the notion of subgamma introduced in Section~2.4 of \citet{boucheron2013concentration} as a key tool for establishing variance concentration:
\begin{definition}
    A real-valued centered random variable $X$ is said to be $(\nu, c)$--subgamma on the right tail with variance factor $\nu$ and scale parameter $c$, denoted as $\Gamma_{+}(\nu, c)$,  if for every $t$ such that $0<t<1/c$:
    \begin{align*}
    \psi_{X}(t) = \log\mathbb{E}[\exp(tX)] \leq \frac{t^{2}\nu}{2(1-ct)},
    \end{align*}
    where $\psi_{X}(t)$ is the cumulant generating function. Similarly, $X$ is said to be $(\nu, c)$-subgamma on the left tail, denoted as $\Gamma_{-}(\nu, c)$, if $–X$ belongs to $\Gamma_{+}(\nu, c)$. If $X$ is $(\nu, c)$-subgamma on both tails, then such random variable is denoted as $\Gamma(\nu,c)$.
\end{definition}

Apart from the subgamma distribution, another closely related and widely used notion is that of subexponential distributions. The theoretical results associated with these two distributional assumptions are largely interchangeable, as many concentration inequalities derived under one setting can be reformulated under the other with comparable bounds. The main distinction lies in the definition: subexponential is typically defined symmetrically for two-sided tails, whereas subgamma treats tails separately. The relationship between subgaussian and subgamma is as follows:

\begin{lemma}\label{lemma:subgamma}
If $X$ follows $\sigma$-subgaussian distribution, then $X^{2}-\mathbb{E}[X^{2}]$ belongs to $\Gamma_{+}(16\sigma^{4}, 2\sigma^{2})$ and $\Gamma_{-}(16\sigma^{4}, \sigma^{2}/3)$. Specifically, if $X$ follows strictly--subgaussian, then $X^{2}-\mathbb{E}[X^{2}]$ belongs to $\Gamma_{+}(2\sigma_{X}^{4},2\sigma_{X}^{2})$ and $\Gamma_{-}(2\sigma_{X}^{4},\sigma_{X}^{2}/3)$.
\end{lemma}

For Lemma~\ref{lemma:subgamma}, if extra information of $X$ is available, such as symmetry or bounded, then we can get more accurate value of $\nu$ and $c$. For example, if $X$ is both symmetric and strictly--subgaussian, then for all $t \in \mathbb{R}$:
    \begin{align*}
        \psi_{X^{2}-\mathbb{E}[X^{2}]}(t) \leq \frac{\sigma_{X}^{4}t^{2}}{1-2\sigma_{X}^{2}t},
    \end{align*}
    which means $X^{2}-\mathbb{E}[X^{2}]$ belongs to $\Gamma_{+}(2\sigma_{X}^{4},2\sigma_{X}^{2})$ and $\Gamma_{-}(2\sigma_{X}^{4},0)$. One typical case is $X \sim \mathcal{N}(0,\sigma_{X}^{2})$.

\subsection{Subgaussian Variance Concentration}
We first revisit several commonly used concentration inequalities relevant to the sample variance. For clarity, let the sample variance of $\{X_{1},\cdots, X_{n}\}$ is defined as:
\begin{align}\label{eq:sample_variance}
    \hat{\sigma}_{n}^{2} = \frac{1}{n-1}\sum_{i=1}^{n}(X_{i}-\bar{X})^2
\end{align}
We assume $\mu_X = 0$ for simplicity, in the case where $\mu_X \neq 0$, the expression can be equivalently rewritten by replacing $X_i$ with $X_i - \mu_X$ in Equation~\eqref{eq:sample_variance}. For $X$ is bounded random variable, one of the famous concentration inequality is Theorem~\ref{thm:bound_concentrate}:

\begin{theorem}(Theorem 10 in \cite{maurer2009empirical}) \label{thm:bound_concentrate}
Let $\{X_{1},\cdots, X_{n}\}$ be $n \geq 2$ i.i.d. random variables with variance $\sigma_{X}^2$ and such that $\{X_{i}\}_{i=1}^{n} \in [0, b]$. Then with probability at least $1-2\delta$, we have:
\begin{align*}
    \left|\hat{\sigma}_{n} - \sigma_{X} \right| \leq b\sqrt{\frac{2\log(1/\delta)}{n-1}}.
\end{align*}
\end{theorem} 

\begin{theorem}(Lemma 4 in \cite{carpentier2011upper}, rephrased) \label{thm:bern}
 Let $X$ follows $\sigma$--subgaussian and variance $\sigma_{X}^{2}$, then with probability at least $1-2\delta$, we have: 
    \begin{align}\label{eq:concentration_baseline}
       |\hat\sigma_{n}-\sigma_{X}| \leq 2\sigma\sqrt{2\log(1/\delta)} \sqrt{\frac{2\log(2/\delta)}{n-1}}+\frac{2\sigma\sqrt{\delta(2-\log(\delta))}}{1-\delta}.
    \end{align}
\end{theorem} 

Theorem~\ref{thm:bern} is an extension of Theorem~\ref{thm:bound_concentrate} under unbounded case, derived through a truncation-based
technique at $|X| \leq \sqrt{2\sigma^{2}\log(1/\delta)}$. However, it could be optimized by utilizing the notion of subgamma and Hanson--Wright inequality. This inequality is particularly powerful in controlling the tail behavior of second-order chaos variables, and is widely applied in analyzing the concentration of quadratic forms.

\begin{theorem}(Theorem~6.2.1 in \cite{vershynin2018hdp}, \cite{epperly2022hanson})\label{thm:hanson_wright}
Let $\boldsymbol X\in\mathbb R^{n}$ have independent, mean-zero, $\sigma$-subgaussian coordinates and $A\in\mathbb R^{n\times n}$ be symmetric. Let $\|\cdot\|_{F}$ denote the Frobenius norm, and $\|\cdot\|$ denote the operator norm, then for all $s\ge0$:
\begin{align}\label{eq:hanson_wright}
    \mathbb{P}\left(|\boldsymbol X^{\top}A\boldsymbol X
        -\mathbb E[\boldsymbol X^{\top}A\boldsymbol X]|\ge s \right) \leq 2\exp\Bigl(
    -\frac{s^{2}/2}{40\sigma^{4}\|A\|_{F}^{2}+8s\sigma^{2}\|A\|}\Bigr).
\end{align}
\end{theorem}

The main proof idea is to separate $A$ as a diagonal matrix $D$ and a diagonal--free matrix $F$, derive the upper bound of $\psi(t)$ of $\boldsymbol X^{\top}D\boldsymbol X$ and $\boldsymbol X^{\top}F\boldsymbol X$ separately, and use Cauchy--Schwartz inequality as:
\begin{align}\label{eq:cauchy-schwarz}
    \psi_{Y+Z}(t) \leq \frac{1}{2}\psi_{Y}(2t)+\frac{1}{2}\psi_{Z}(2t).
\end{align}
Theorem~\ref{thm:hanson_wright} provides a general concentration result for any symmetric matrix \( A \). In the special case of the sample variance, we can express it as $\hat{\sigma}_{n}^{2} = \boldsymbol X^{\top}A\boldsymbol X$ with $A = \tfrac{1}{n-1}(I_n-\tfrac1n\mathbf1\mathbf1^{\top})$. Leveraging this specific structure, we can apply H\"older's inequality in equation~\eqref{eq:cauchy-schwarz} to derive a more refined concentration bound of the sample variance estimator, which is a tighter version of Lemma~\ref{lemma:variance_concentrarion}:
\begin{corollary}(A tighter version of Lemma~\ref{lemma:variance_concentrarion}) \label{corollary:concentration_variance}
Suppose $X$ is $\sigma$-subgaussian, then we have:
\begin{align*} 
    \mathbb{P}\left(\hat\sigma_n^{2}-\sigma_X^{2}\ge 4\sigma^{2}f(n)\sqrt{\frac{2\log(1/\delta)}{n-1}}+\frac{6\sigma^{2}\log(1/\delta) }{n}\right)
    \le \delta, \\ \mathbb{P}\left(\sigma_X^{2}-\hat\sigma_n^{2} \ge 4\sigma^{2}f(n)\sqrt{\frac{2\log(1/\delta)}{n-1}}+\frac{13\sigma^{2}\log(1/\delta)}{3n}\right) \le \delta.
\end{align*}
where $f(n) = (1+\sqrt{n-1})/\sqrt{n}$. Specifically, if $X$ is strictly subgaussian, then $f(n) = (1+\sqrt{(n-1)/8})/\sqrt{n}$, and $\sigma$ can be replaced with $\sigma_{X}$. 
\end{corollary}

The detailed proof is in Appendix~\ref{appendix:proof_lemma1}. For Lemma~\ref{lemma:variance_concentrarion}, a slightly weaker but still informative version is:
    \begin{align*}
        \mathbb{P}\big(|\hat\sigma_n^{2}-\sigma_X^{2}|\ge s\bigr) \leq 2\exp(
    -\frac{s^{2}/2}{32\sigma^{4}/(n-1)+6s\sigma^{2}/n}), \ \text{for } s>0.
    \end{align*}
Using the fact that $\|A\|_{F}^{2} = \|A\|_{2} = 1/(n-1)$ for the sample variance, this bound is tighter than the general bound~\eqref{eq:hanson_wright}. Additionally, using the inequality $|\hat{\sigma}_{n}-\sigma_{X}|\;\leq\;\frac{|\hat{\sigma}^{2}_{n}-\sigma^{2}_{X}|}{\sigma_{X}}$, with probability at least $1-2\delta$, we have:
\begin{align*}
    |\hat{\sigma}_{n}-\sigma_{X}|
    \;\leq\; \frac{4\sigma^{2}(1+\sqrt{n-1})}{\sigma_{X}\sqrt{n}}\,
    \sqrt{\frac{2\log(1/\delta)}{n-1}}
    +\frac{6\sigma^{2}\log(1/\delta)}{n}.
\end{align*}
Compared with the bound~\eqref{eq:concentration_baseline}, this result removes one factor of $\sqrt{\log(1/\delta)}$. In particular, when setting $\delta = T^{-\alpha_{0}}$ with $\alpha_{0}\geq 1$, as is commonly done in bandit problems, the improvement amounts to eliminating one $\sqrt{\log T}$ term.

\section{More Detailed Theoretical Analyses}\label{appendix:theoretical_supp}

In this section, we expand on the theoretical analyses of our algorithms from Sections~3 and~4, which were only briefly outlined in the main text due to space constraints. In addition, we specify the setting corresponding to a strictly subgaussian distribution. We use $\varepsilon^{-}_{n}(\delta)$ and $\varepsilon^{+}_{n}(\delta)$ represent the confidence bound shown in Corollary~\ref{corollary:concentration_variance}. This notation could be simplified as $\varepsilon_{n}(\delta)$ if using Lemma~\ref{lemma:variance_concentrarion}. 

In our framework, the algorithms select each $k$-th alternative for $\tau_k$ rounds ($\tau_k = \tau$ in the non-adaptive case), then estimates $\lambda_{k,\tau_k}$ for each arm, and subsequently allocates the remaining rounds according to these estimates. If we ensure that $\tau_k \le n_k^{}$ for all $k = 1, \ldots, K$, we can define
\begin{align*}
    \bm{\lambda} &\coloneqq \bm{n}/T = \{\lambda_{1,\tau_{1}},\cdots, \lambda_{K,\tau_{K}}\}, \\
    \bm{\lambda}^{*} &\coloneqq \bm{n}^{*}/T = \{\lambda^{*}_{1},\cdots, \lambda^{*}_{K}\}, 
\end{align*}
then the objective function can be expressed equivalently as:
\begin{equation*}
R_{p}(\boldsymbol{n}) = R_p(\bm{\lambda}) = \frac{1}{T}
\begin{cases}
\bigg(\displaystyle\sum_{k=1}^{K}\Big(\frac{\sigma_k^2}{\lambda_k}\Big)^{p}\bigg)^{\!1/p}, & p<\infty,\\[6pt]
\displaystyle\max_{1\le k\le K}\frac{\sigma_k^{2}}{\lambda_k}, & p=\infty.
\end{cases}
\end{equation*}

\subsection{Detailed Analyses for Section~3}\label{appendix:section3}
Since in Algorithm~\ref{alg:FETE}, each alternative is chosen for same times, and the concentration bound at time $n$ is same for each alternative, then we can use $\varepsilon^{-}_{k,n}(\delta) = \varepsilon^{-}_{n}(\delta)$ and $\varepsilon^{+}_{k,n}(\delta) = \varepsilon^{+}_{n}(\delta)$ to denote the concentration bound for $k$-th alternative for simplicity. Since Algorithm~\ref{alg:FETE} ensures that
$\tau \le \min_{k=1,...,K} n_k^*$, then we have $\lambda_{k,\tau_{k}} = \lambda_{k,\tau}$. 
To give a high-probability theoretical bound, as Algorithm~\ref{alg:FETE} is non-adaptive, let $\delta = T^{-1}$ for infinite case and $\delta = T^{-3/2}$ for finite case, and define the event
\begin{align*}
    \xi_{\tau}(\delta) = \{-\varepsilon^{-}_{\tau}(\delta) \leq \hat{\sigma}^{2}_{k,\tau}-\sigma^{2}_{k}\leq \varepsilon^{+}_{\tau}(\delta) \text{ for all }k\},
\end{align*}
then we have $\mathbb{P}(\xi^{c}_{\tau}(\delta)) \leq 2K\delta$. We begin by considering the case \( p = \infty \), where the regret difference simplifies to:
\begin{align*}
    R_{\infty}(\bm{\lambda}) - R_{\infty}(\bm{\lambda^{*}}) 
    = \frac{1}{T} \left( \max_{1 \leq k \leq K} \frac{\sigma_k^2}{\lambda_{k,\tau}} - \Sigma_2 \right).
\end{align*}

Suppose $\xi_{\tau}(\delta)$ exits, since now the confidence radius is identical across all arms, the worst-case scenario occurs when the arm with the smallest true variance realizes its lower confidence bound, while all other arms attain their respective upper bounds. Let $\bm{\lambda}_{\pi_1} = \bm{n}_{\pi_1}\cdot T^{-1}$, then we could achieve Theorem~\ref{thm:pinf} based on Lemma~\ref{lemma:Rgapinf}.

\begin{lemma} \label{lemma:Rgapinf}
    If $\xi_{\tau}(\delta)$ exists and
    $\varepsilon^{-}_{\tau}(\delta) < \sigma^{2}_{\min}$, then we have:
    \begin{align*}
         R_{\infty}(\bm{\lambda}_{\pi_1}) - R_{\infty}(\bm{\lambda^{*}}) \leq \frac{1}{T}\bigg[(K-1)\varepsilon^+_{\tau}+(\frac{\Sigma_{2}}{\sigma^{2}_{\min}}-1)\varepsilon^{-}_{\tau}\bigg]+o(T^{-3/2}).
    \end{align*} 
\end{lemma}

When \( p \) is finite, we could leveraging the smoothness property of \( R_{p}(\bm{\lambda}) \) based on Lemma~\ref{lemma:Rgaptaylor}, which is Lemma~4 in \cite{aznag2023an}. But as it is difficult to identify the worst-case configuration of the variance estimates, we give a slightly loose upper bound Theorem~\ref{thm:pfinite} based on Lemma~\ref{lemma:Rgapfinite}.

\begin{lemma} (Lemma 4 in \cite{aznag2023an}, rephrased)\label{lemma:Rgaptaylor}
    If $p$ is finite, then we have:
    \begin{align*}
        R_{p}(\bm{\lambda}) - R_{p}(\bm{\lambda}^{*}) \leq \frac{(p+1)R_{p}(\bm{\lambda}^{*})}{2}\sum_{k=1}^{K}\frac{(\lambda_{k}-\lambda^{*}_{k})^{2}}{\lambda^{*}_{k}} + \frac{7(p+2)^{2}}{\lambda_{\min}^{*}T}\max\limits_{k}(\frac{\lambda_{k}^{*}}{\lambda_{k}})^{3p+3}||\bm{\lambda} - \bm{\lambda}^{*}||^3_{\infty}.
    \end{align*}
\end{lemma}

\begin{lemma}\label{lemma:Rgapfinite}
    If $\xi_{\tau}(\delta)$ exists and
    $\varepsilon^{-}_{\tau}(\delta) < \sigma^{2}_{\min}$, then for $p$ is finite, we have:
    \begin{align*}
         R_{p}(\bm{\lambda}_{\pi_1}) - R_{p}(\bm{\lambda}^{*}) \leq \frac{p^{2}(\Sigma_{q})^{1/p}\Sigma_{q-4}}{2(p+1)T}(\varepsilon^{+}_{\tau})^{2}+o(T^{-2}). 
    \end{align*}
\end{lemma}

\subsection{Detailed discussion in Section 4}\label{appendix:section4}
We begin with introducing the design of LCBs and UCBs in Section~4. Let $\delta = T^{-2}$ for infinite case and $\delta = T^{-5/2}$ for finite case, and define the event \( \xi_{T}(\delta) \) as:
\begin{align*}
    \xi_{T}(\delta) = \left\{ -\varepsilon^{-}_{k,n}(\delta) \leq \hat{\sigma}^{2}_{k,n} - \sigma^{2}_{k} \leq \varepsilon^{+}_{k,n}(\delta) \text{ for all }k\text{ and }n \right\},
\end{align*}
then we have
$\mathbb{P}(\xi^{c}_{T}(\delta)) \leq 2T\delta$.  For each arm \( k \) and time step \( n \), for general subgaussian case, define the lower and upper confidence bounds for arm \( k \) at time \( n \) as:
\[
\text{LCB}_{k,n} = \max\{\hat{\sigma}^{2}_{k,n} - \varepsilon^{+}_{k,n},0\}, \quad \text{UCB}_{k,n} = \hat{\sigma}^{2}_{k,n} +\varepsilon^{-}_{k,n},
\]
then we can get $$\mathbb{P}\left(\text{LCB}_{k,n}\le \sigma_k^2 \le \text{UCB}_{k,n}\text{ for all }k\text{ and }n \right) = \mathbb{P}(\xi_{\tau}(\delta)) \ge1 - 2T^{-c},$$
with $c = 1$ for $p=\infty$ and $c=3/2$ for $p<\infty$.

For LCBs, we need to make sure $\min_{k}\text{LCB}_{k,n} >0$ otherwise the second phase of Algorithm~\ref{alg:adaptiveETC} would fail to work. To avoid confusion, let $\alpha_{k} \leq \tau_{k}/T$ at the end of Phase~2. Building on the analysis presented before, the key challenge in the general setting is to identify $\alpha_{k}$ such that $\alpha_{k}T \leq \lambda^{*}_{k}T$. Here we consider the general subgaussian case, the following lemma characterizes the corresponding relationship between \( \alpha_{k}\) and $\lambda^{*}_{k}$ when $T$ is large:
\begin{lemma}\label{lemma:general_alpha}
    If \( \xi_{T}(\delta) \) holds, then the exploration length \( \tau_{k} \geq \alpha_{k} T\), with $\alpha_{k}  = \lambda^{*}_{k}(1-\Theta(\sqrt{T^{-1}\log T}))$.
\end{lemma}

Then based on Lemma~\ref{lemma:general_alpha}, and follow the proof procedure of Theorem~\ref{thm:pinf} and~\ref{thm:pfinite}, we could finally achieve Theorem~\ref{thm:regretAlg3_subgaussian}. In Theorem~\ref{thm:regretAlg3_subgaussian} when $p=\infty$, careful readers may notice additional factors such as $\Sigma_{-1}$ and $\sigma_{\min}^{-1}$ in our bound compared with the result in \citet{aznag2023an}. These terms stem from our exploration-free design: the estimate of $\lambda_{k,\tau_k}$ is computed only after each arm has been pulled $\tau_k$ times, and $\tau_k$ is close to $\lambda_{\min}^* T$ by Lemma~\ref{lemma:general_alpha}. In the worst-case regret, we therefore include $\varepsilon_{\tau_k}$, whose leading term scales as $\sqrt{\log T / \tau_k}$, which in turn yields the factors $\Sigma_{-1}$ and $\sigma_{\min}^{-1}$. However, we need to stress that our Algorithm~\ref{alg:adaptiveETC} removes one factor of $\log T$ relative to \citet{aznag2023an} beyond the refined concentration inequality, resulting in a strictly tighter asymptotic rate.

If prior information of $\underline{\sigma}^{2}$ is available, it still can be effectively incorporated into Algorithm~\ref{alg:adaptiveETC}. Rather than playing each arm twice, one may directly allocate $\tau$ rounds as prescribed by Equation~\eqref{eq:tau_nonadaptive} to each arm in the first phase, since this value serves as a deterministic lower bound for sufficient exploration. 

\subsection{Strictly--Subgaussian Case Analysis}\label{appendix:strictlysubgaussian}
For the strctly--subgaussian case, based on Lemma~\ref{lemma:variance_concentrarion} we know that the concentration bound is proportional to the real variance for each arm $k$, then we can use $\varepsilon^{-}_{k,n}(\delta) = \sigma^{2}_{k}\cdot s^{-}_{n}(\delta)$ and $\varepsilon^{+}_{k,n}(\delta) = \sigma^{2}_{k}\cdot s^{+}_{n}(\delta)$ to represent the concentration bound of arm $k$ for simplicity, which means for  $\xi_{\tau}(\delta)$ in Algorithm~\ref{alg:FETE}, we have:
\begin{align*}
    \sigma^{2}_{k}\cdot (1-s^{-}_{\tau}(\delta)) \leq \hat{\sigma}^{2}_{k,\tau}\leq \sigma^{2}_{k}\cdot (1+s^{+}_{\tau}(\delta))
\end{align*}
Then we can get the following result based on Theorem~\ref{thm:pinf} and Theorem~\ref{thm:pfinite}:

\begin{theorem}\label{thm:strictly_p}
    In the strictly--subgaussian setting of Algorithm \ref{alg:FETE}, for $p=\infty$, let $\delta = T^{-1}$, we have:
    \begin{align*}
        \mathbb{E}\left[R_{p}(\bm{n}_{\pi_1}) - R_{p}(\bm{n}^{*})\right] &\leq 4\lambda^{-1/2}(\Sigma_{2}-\sigma^{2}_{\min})
        T^{-3/2}\sqrt{\log T}+o(T^{-3/2}).
    \end{align*}
    For $p$ is finite, let $\delta = T^{-3/2}$, we have:
    \begin{align*}
        \mathbb{E}\left[R_{p}(\bm{n}_{\pi_1}) - R_{p}(\bm{n}^{*})\right] \leq \frac{3p^{2}(\Sigma_{q})^{2/q}}{\lambda(p+1)}T^{-2}\log T+o(T^{-2}).
    \end{align*}
    where $\lambda$ is defined in Section~3.1.
\end{theorem}

For Algorithm~\ref{alg:adaptiveETC} as an adaptive process, then we could define the lower and upper confidence bound for each arm $k$ at time $n$ as:
\begin{align*}
    \text{LCB}_{k,n} = \frac{\hat{\sigma}^{2}_{k,n}}{1+s^{+}_{n}}, \quad \text{UCB}_{k,n} = \frac{\hat{\sigma}^{2}_{k,n}}{1-s^{-}_{n}}.
\end{align*}
Then in the event $\xi_{T}(\delta)$, we have:
\begin{align*}
    \sigma^{2}_{k}\cdot \frac{1-s^{-}_{n}}{1+s^{+}_{n}} \leq \text{LCB}_{k,n} \leq \sigma^{2}_{k} \leq \text{UCB}_{k,n} \leq \sigma^{2}_{k}\cdot \frac{1+s^{+}_{n}}{1-s^{-}_{n}}.
\end{align*}

The particular form of our LCB and UCB has a convenient property in the first phase of Algorithm~\ref{alg:adaptiveETC}: it suffices to ensure
$$s^{-}_{n}(\delta) < 1.$$
Because $s^{-}_{n}(\delta)$ depends only on $(n,\delta)$, once $\delta$ is fixed we can compute the exact minimal $n$ required to meet this condition, independently of $\sigma^{2}$ and $\sigma_{\min}^{2}$. This decoupling not only simplifies the analysis and implementation but can also reduce the number of initial samples needed. For the theoretical analysis, based on Lemma~\ref{lemma:variance_concentrarion} and~\ref{lemma:general_alpha}, it is easy to certify that Lemma~\ref{lemma:general_alpha} would still be satisfied in strictly--subgaussian case. Then we have the following result:

\begin{theorem}\label{thm:regretAlg3_strictlysubgaussian}
In the strictly--subgaussian setting of Algorithm~\ref{alg:adaptiveETC}, for $p=\infty$, let $\delta = T^{-2}$, we have:
    \begin{align*}
        \mathbb{E}\left[R_{p}(\bm{n}_{\pi_2}) - R_{p}(\bm{n}^{*})\right] &\leq 2\sqrt{2}\left[\frac{\sqrt{\Sigma_{2}}(\Sigma_{2}-2\sigma^{2}_{\min})}{\sigma_{\min}}+\sqrt{\Sigma_{2}}\Sigma_{1}\right]T^{-3/2}\sqrt{\log T}+o(T^{-3/2}). 
    \end{align*}
For $p$ is finite, let $\delta = T^{-5/2}$, we have:
    \begin{align*}
        \mathbb{E}\left[R_{p}(\bm{n}_{\pi_2}) - R_{p}(\bm{n}^{*})\right] &\leq \frac{5Kp^{2}(\Sigma_{q})^{2/q}}{p+1}T^{-2}\log T+o(T^{-2}).
    \end{align*}
\end{theorem}

As established in Section~2.1, the leading-order terms in the concentration bounds are equivalent for strictly--subgaussian and Gaussian distributions. Consequently, the regret result in the strictly--subgaussian setting remains valid for the Gaussian case, with the only modification being the substitution of $s^{+}_{n}$ and  $s^{-}_{n}$ with their counterparts derived from the Gaussian-specific concentration bounds in Equation~\eqref{eq:gaussian_concentration}. The  difference of the results between these two cases lies solely in an asymptotically negligible residual term. Then based on Theorem~\ref{thm:regretAlg3_strictlysubgaussian}, we could also get a tighter version for Theorem~\ref{thm:Contextual_regret} when the variances belong to strictly- subgaussian distribution:
\begin{theorem}\label{thm:Contextual_regret_SSG}
    In the strictly-subgaussian setting of Algorithm \ref{alg:contextualETC}, with $p=1$, we have 
    \begin{align*}
        \mathbb{E}[R_T(\bm{n}_{\pi_{3}}) - R_T(\bm{n}^{*})] \leq \frac{5dK(\Sigma_{1})^{2}}{\lambda^{\mathcal{C}}_{\min}}T^{-2}\log T+o(T^{-2}).
    \end{align*}
\end{theorem}

\subsection{Upper Confidence Bound in Phase~3}
For $p=\infty$, if we use an UCB estimator in the third phase of Algorithm~\ref{alg:adaptiveETC} to calculate $\lambda_{k,\tau_k}$ as:
\begin{align*}
    \lambda_{k,\tau_{k}} = \frac{(\hat{\sigma}^{2}_{k,\tau_{k}}+\varepsilon^{-}_{k,\tau_{k}})^{q/2}}{(\hat{\sigma}^2_{k,\tau_{k}}+\varepsilon^{-}_{k,\tau_{k}})^{q/2} +\sum_{j\neq k}(\hat{\sigma}^{2}_{j,\tau_{j}}+\varepsilon^{-}_{j,\tau_{j}})^{q/2}},
\end{align*}
then for the general subgasusian setting, we can get a tighter value of $\FalgInf{2}{\bm{\sigma^{2}}}$:
$$\FalgInf{2}{\bm{\sigma^{2}}} = 2\sqrt{\Sigma_{2}}(\Sigma_{-1}-\sigma^{-1}_{\min}),$$
and for the strictly subgaussian setting, we would have:
\begin{align*}
    \mathbb{E}\left[R_{p}(\bm{n}_{\pi_2}) - R_{p}(\bm{n}^{*})\right] &\leq 4\sqrt{2}\left[\sqrt{\Sigma_{2}}(\Sigma_{1}-\sigma_{\min})\right]T^{-3/2}\sqrt{\log T}+o(T^{-3/2}). 
\end{align*}

\section{More Experiments and Experiment Details}\label{appendix:experiments}
In this section, we provide some experiment details and present some additional numerical experiments.

\subsection{Experiment Details}
In our numerical experiments, the initial run length per arm in the first phase of Algorithm~\ref{alg:adaptiveETC} when $\underline{\sigma}^{2}$ is unknown is chosen as $\min\{64 \sigma^{4}\log T, T/K\}$ for the general subgaussian setting and $\min\{18\log T, T/K \}$ for the strictly subgaussian setting.
When $\underline{\sigma}^{2}$ is known, we set $\tau$ as in~\eqref{eq:tau_nonadaptive}. In all cases, we then increment the number by one until the first phase stopping condition is satisfied.

Besides, in the second phase of Algorithm~\ref{alg:adaptiveETC}, instead of checking the condition \(n_k \ge \lambda_{k,n_k}\,T\) after every unit increase of \(n\), we use an iterative batched thresholding scheme, which reduces the number of feasibility tests and improves running time. We need to note that per-increment checks might yield slightly finer stopping times, and thus marginally more accurate empirical allocations. Here the batched scheme achieves nearly identical outcomes in practice at a substantially lower computational cost.

Finally, since we apply the floor function at each allocation update to enforce integer sample counts, which may leave a small residual budget due to rounding. After the main allocation, we greedily top up by assigning the leftover rounds sequentially to the arms with the largest estimated variances. This final correction ensures $\sum_{k} n_k = T$ while preserving the scheme’s asymptotic optimality, since the correction is at most $K$ times and thus negligible relative to $T$.

\subsection{Traditional MAB: Gaussian arms}\label{appendix:experiments_Gaussian}
\begin{figure}[htbp]
  \centering
  \begin{subfigure}{0.32\columnwidth}
    \centering
    \includegraphics[width=\linewidth]{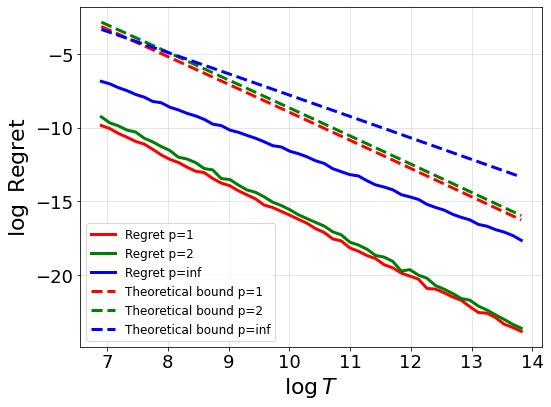}
    \caption{GSG with $\underline{\sigma}^2$}
  \end{subfigure}\hfill
  \begin{subfigure}{0.32\columnwidth}
    \centering
    \includegraphics[width=\linewidth]{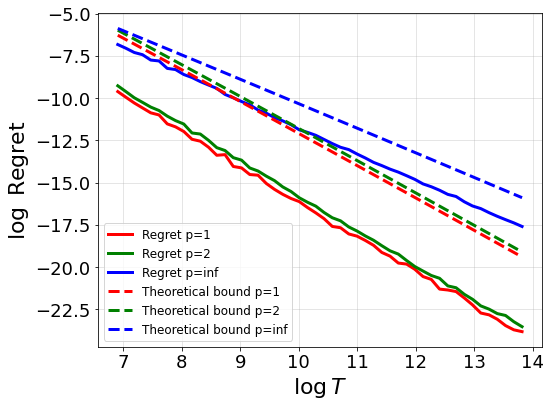}
    \caption{SSG with $\underline{\sigma}^2$}
  \end{subfigure}\hfill
  \begin{subfigure}{0.32\columnwidth}
    \centering
    \includegraphics[width=\linewidth]{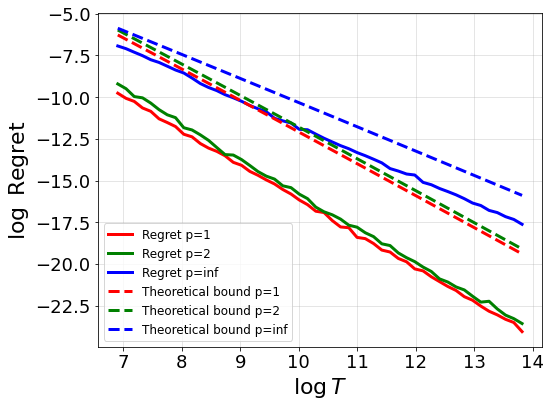}
    \caption{Gaussian with $\underline{\sigma}^2$}
  \end{subfigure}

  \medskip

  \begin{subfigure}{0.32\columnwidth}
    \centering
    \includegraphics[width=\linewidth]{Figures/Numerical_generalSG_unknown.png}
    \caption{GSG without $\underline{\sigma}^2$}
  \end{subfigure}\hfill
  \begin{subfigure}{0.32\columnwidth}
    \centering
    \includegraphics[width=\linewidth]{Figures/Numerical_SSG_unknown.png}
    \caption{SSG without $\underline{\sigma}^2$}
  \end{subfigure}\hfill
  \begin{subfigure}{0.32\columnwidth}
    \centering
    \includegraphics[width=\linewidth]{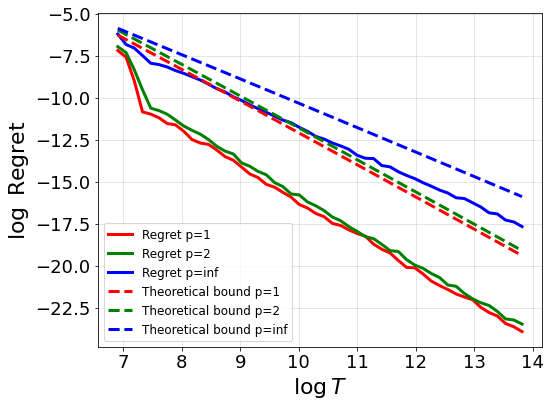}
    \caption{Gaussian without $\underline{\sigma}^2$}
  \end{subfigure}

  \caption{Algorithm~\ref{alg:adaptiveETC}: comparison of three regimes with and without $\underline{\sigma}^2$.}
  \label{fig:three_regimes}
\end{figure}

Here we provide more numerical figures as in our first Gaussian alternative experiment. From Figure~\ref{fig:three_regimes}, we first observe that the performance of SSG closely resembles that of the Gaussian case, indicating that SSG effectively captures the key properties of Gaussian noise. By contrast, when $\underline{\sigma}^{2}$ is known, the second phase can be entered much earlier, since $\underline{\sigma}^{2}$ provides a lower bound for $n^{*}_{\min}$ and thereby circumvents the stringent exploration requirement in the first phase.

\subsection{Another Strictly Subgaussian Example: Symmetric Beta}
To further validate the benefits of the strictly--subgaussian property, we consider a non-Gaussian example based on the symmetric Beta distribution $\mathrm{Beta}(\alpha,\alpha)$, supported on $[0,1]$. This family is flexible: for $\alpha<1$ it is U-shaped, while larger $\alpha$ values concentrate mass near $0.5$, with variance $(2\alpha+1)^{-1}$, and it is known to satisfy strictly--subgaussian properties, making it a natural candidate for studying group mean estimation.  

In our experiment, we simulate $K=4$ arms with rewards $X_{k,n}=\mu_k+\mathrm{Beta}(\alpha,\alpha)$, where $\mu_k\sim\mathcal{U}([-1,1])$. The shape parameters $\{\alpha_k\}=\{0.2,1.0,2.0,4.5\}$ are chosen to cover a wide spectrum of tail behaviors. We set $\underline{\sigma}^2 = \sigma^{2}_{\min}$ and $\sigma^{2}=1$ from the variance formula. 

\begin{figure}[htbp]
  \centering
  \begin{subfigure}[t]{0.45\columnwidth}
    \centering
    \includegraphics[width=\linewidth]{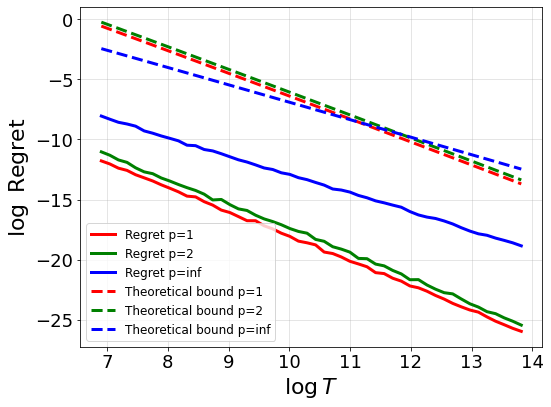}
    \caption{GSG Setting with known $\underline{\sigma}$}
  \end{subfigure}\hfill
  \begin{subfigure}[t]{0.45\columnwidth}
    \centering
    \includegraphics[width=\linewidth]{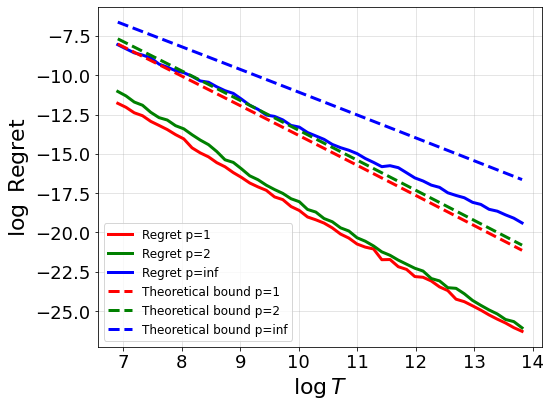}
    \caption{SSG Setting with known $\underline{\sigma}$}
  \end{subfigure}\hfill
  \caption{Algorithm~\ref{alg:adaptiveETC}: Symmetric Beta.}
  \label{fig:Strictly_SUBGAUSSIAN_beta}
\end{figure}

\section{Proof of Section~2}
This section gives the proof of Proposition~\ref{prop:optimal} and Lemma~\ref{lemma:variance_concentrarion} in Section~2.

\subsection{Proof of Proposition~\ref{prop:optimal}}
\begin{proof}
For the case $p<\infty$, since $f(x) =  x^{1/p}$ is non-decreasing, it is equivalent to minimizing $\sum_{k=1}^K \sigma_k^{2p}/n_k^{p}$ under $\sum_{k=1}^{K} n_k=T$. By using the Lagrangian and the first-order optimality conditions, we can get that $n_k^{p+1}\propto \sigma_k^{2p}$, therefore we have:
\begin{align*}
    n_k^{*}=\frac{\sigma_k^{\frac{2p}{p+1}}}{\sum_{j=1}^K \sigma_j^{\,\frac{2p}{p+1}}}T
    =\frac{\sigma_k^{q}}{\sum_{j=1}^K \sigma_j^{q}}T, 
\end{align*}
then plug into $R_p$, we have: 
\begin{align*}
    R_p(\bm n^*)=\left[\Big(\tfrac{\sum_j \sigma_j^{q}}{T}\Big)^p \sum_k \sigma_k^{q}\right]^{1/p}
    =\frac{1}{T}\bigg(\sum_{k=1}^K \sigma_k^{q}\bigg)^{\frac{p+1}{p}}
    =\frac{1}{T}\bigg(\sum_{k=1}^K \sigma_k^{q}\bigg)^{\frac{2}{q}}.    
\end{align*}

For the case $p=\infty$, the objective is equivalent to
minimize $t$ such that \ $\sigma_k^2/n_k\le t$ for all $k$ and $\sum_{k=1}^{K} n_k=T$. Then we can get $n_k\ge \sigma_k^2/t$, hence $t \ge \sum_{k=1}^{K} \sigma_k^2/T$. Then we could get that:
\begin{align*}
    n_k^* = \frac{\sigma_k^2}{\sum_{j=1}^{K} \sigma_j^2} T, \quad R_\infty(\bm{n}^*)=\frac{1}{T}\sum_{k=1}^K \sigma_k^2.
\end{align*}
\end{proof}   

\subsection{Proof of Lemma~\ref{lemma:variance_concentrarion}}\label{appendix:proof_lemma1}

\begin{proof}
The proof is based on Section~2 of \citet{boucheron2013concentration}, Section~6 of \citet{vershynin2018hdp} and the theoretical framework of \citet{epperly2022hanson}. For the sample variance, let $D = \frac{1}{n}I_{n}$ and $F=-\frac{1}{n(n-1)}\mathbf1\mathbf1^{\top}+\frac{1}{n(n-1)}I_{n}$, then we have $\hat{\sigma}_{n}^{2} = \boldsymbol X^{\top}A\boldsymbol X =  \boldsymbol X^{\top}D\boldsymbol X+\boldsymbol X^{\top}F\boldsymbol X$ and $\mathbb{E}[\boldsymbol X^{\top}A\boldsymbol X] =  \mathbb{E}[\boldsymbol X^{\top}D\boldsymbol X]$.

We first show the right-tail bound when $X$ belongs to $\sigma$--subgaussian. For the diagonal matrix $D$, since each entry $d_{i,i} = n^{-1}$, then for $t>0$, based on Lemma~\ref{lemma:subgamma}, we have:
\begin{align*}
    \psi_{\boldsymbol X^{\top}D\boldsymbol X-\mathbb{E}[\boldsymbol X^{\top}A\boldsymbol X]}(t) =\frac{8\sigma^{4}t^{2}n^{-1}}{1-2\sigma^{2} t/n}
\end{align*}
For the diagonal--free matrix $F$, we apply the decoupling bound together with the standard Gaussian quadratic form representation $\tilde{g}^{\top}Fg$ (see Theorem~6.1.1 and Lemma~6.2.3 in \citet{vershynin2018hdp}, \citet{epperly2022hanson}) to obtain:
\begin{align*}
     \psi_{\boldsymbol X^{\top}F\boldsymbol X}(t) 
    \leq \psi_{\tilde{g}^{\top}Fg}(4\sigma^{2}t) \leq  
    \frac{8\sigma^{4}\|F\|^{2}_{F}t^{2}}{1-4\sigma^{2}\|F\|t} = \frac{8\sigma^{4}t^{2}/(n^{2}-n)}{1-4\sigma^{2}t/n},
\end{align*}
where the second inequality is by Hermitian dilation: $\psi_{\tilde{g}^{\top}Fg}(t) \leq  \frac{\|F\|^{2}_{F}t^{2}}{2(1-\|F\|t)}$, the last equality is from $\|F\|_{F}^{2}  = 1/(n^{2}-n)$ and $\|F\|  = 1/n$. Then we can apply H\"older's inequality, let positive $p$, $q$ satisfy $1/p+1/q=1$:
\begin{align*}
    \psi_{\boldsymbol X^{\top}A\boldsymbol X-\mathbb{E}[\boldsymbol X^{\top}A\boldsymbol X]}(t) &\leq \frac{1}{p}\psi_{\boldsymbol X^{\top}D\boldsymbol X-\mathbb{E}[\boldsymbol X^{\top}A\boldsymbol X]}(pt)+\frac{1}{q}\psi_{\boldsymbol X^{\top}F\boldsymbol X}(qt)\\
    &=\frac{8\sigma^{4} pt^{2}n^{-1}}{1-2\sigma^{2} pt/n}+\frac{8\sigma^{4} qt^{2}/(n^{2}-n)}{1-4\sigma^{2}qt/n}\\
    &= 8\sigma^{4}t^{2}\bigg[\frac{1}{n/p-2t\sigma^{2}}+\frac{1}{(n-1)[(1-1/p)n-4\sigma^{2} t]}\bigg]\\
    &= \frac{8\sigma^{4} t^{2}(1+\sqrt{n-1})^{2}/(n^{2}-n)}{1-6\sigma^{2} t/n}
\end{align*}
where for the last equality, let $f(x) = \frac{1}{nx-a}+\frac{C}{(n-1)[(1-x)n-b]}$, after some elementary calculation, we can get that when $n > a+b$, with $x^{*} = \frac{(n-b)\sqrt{(n-1)/C}+a}{n(1+\sqrt{(n-1)/C})} <1$, $\min f(x) = f(x^{*}) = \frac{(1+\sqrt{C/(n-1)})^{2}}{n-(a+b)} = \frac{C(1+\sqrt{(n-1)/C})^{2}/(n^{2}-n)}{1-(a+b)/n}$, then we can use $1/p = x^{*}$.

This implies $\boldsymbol X^{\top}A\boldsymbol X-\mathbb{E}[\boldsymbol X^{\top}A\boldsymbol X]$ belongs to $\Gamma_{+}(\nu,c)$ with $\nu = 16\sigma^{4}(1+\sqrt{n-1})^{2}/(n^{2}-n)$ and $c = (2+4/(n-1))\sigma^{2}/n$. By Chernoff's inequality (see Section~2.4 in \cite{boucheron2013concentration}), 
if $X\in \Gamma_{+}(\nu,c)$ then for any $s>0$, $\mathbb{P}\!\left(X>\sqrt{2\nu s}+cs\right)\;\leq\; e^{-s}$. Then set $s=\log(1/\delta)$, we can get:
\[
\mathbb{P}\left(\hat\sigma_n^{2}-\sigma_X^{2}\ge \frac{4\sigma^{2}(1+\sqrt{n-1})}{\sqrt{n}}\sqrt{\frac{2\log(1/\delta)}{n-1}}+\frac{6\sigma^{2}\log(1/\delta) }{n}\right)
    \le \delta.
\]

For the left-tail bound, for $t>0$, since $-\boldsymbol X^{\top}A\boldsymbol X+\mathbb{E}[\boldsymbol X^{\top}A\boldsymbol X] = -\boldsymbol X^{\top}D\boldsymbol X+\mathbb{E}[\boldsymbol X^{\top}A\boldsymbol X]+\boldsymbol X^{\top}(-F)\boldsymbol X$, then based on Lemma~\ref{lemma:subgamma}, $X$ belongs to $\Gamma_{-}(16\sigma^{4},\sigma^{2}/3)$, we have:
\begin{align*}
    \psi_{\boldsymbol X^{\top}(-D)\boldsymbol X+\mathbb{E}[\boldsymbol X^{\top}A\boldsymbol X]}(t) \leq \frac{8\sigma^{4}t^{2}}{n-\sigma^{2}t/3}, \quad \psi_{\boldsymbol X^{\top}(-F)\boldsymbol X}(t) \leq \psi_{\tilde{g}^{\top}(-F)g}(4\sigma^{2}t)\leq \frac{8\sigma^{4}t^{2}/(n^{2}-n)}{1-4\sigma^{2}t/n}.
\end{align*}
Then we can write the cumulant generating function as:
\begin{align*}
    \psi_{-\boldsymbol X^{\top}A\boldsymbol X+\mathbb{E}[\boldsymbol X^{\top}A\boldsymbol X]}(t) &\leq \frac{1}{p}\psi_{-\boldsymbol X^{\top}D\boldsymbol X+\mathbb{E}[\boldsymbol X^{\top}A\boldsymbol X]}(pt)+\frac{1}{q}\psi_{\boldsymbol X^{\top}(-F)\boldsymbol X}(qt) \\
    &= \frac{8\sigma^{4}pt^{2}n^{-1}}{1-\sigma^{2}pt/3n}+\frac{8\sigma^{4}t^{2}/(n^{2}-n)}{1-4\sigma^{2}t/n}\\
    &= 8\sigma^{4}t^{2}\left[\frac{1}{n/p-\sigma^{2}t/3}+\frac{1}{(n-1)[(1-1/p)n-4\sigma^{2} t]} \right]\\
    &= \frac{8\sigma^{4}t^{2}(1+\sqrt{n-1})^{2}/(n^{2}-n)}{1-13\sigma^{2}t/3n}
\end{align*}
Then $\boldsymbol X^{\top}A\boldsymbol X-\mathbb{E}[\boldsymbol X^{\top}A\boldsymbol X]$ belongs to $\Gamma_{-}(\nu,c)$ with $\nu = 16\sigma^{4}(1+\sqrt{n-1})^{2}/(n^{2}-n)$, $c = 13\sigma^{2}/3n$. Then we can get the result by using Chernoff's inequality.

For the strcitly--subgaussian case, the proof is similar, now we have $X^{2}-\mathbb{E}[X^{2}]$ belongs to $\Gamma_{+}(2\sigma^{4}_{X},2\sigma_{X}^{2})$ and $\Gamma_{-}(2\sigma^{4}_{X},\sigma_{X}^{2}/3)$ as proved in Lemma~\ref{lemma:subgamma}. Following the proof above, we can get that $\boldsymbol X^{\top}A\boldsymbol X-\mathbb{E}[\boldsymbol X^{\top}A\boldsymbol X]$ belongs to $\Gamma_{+}(\nu ,c^{+})$ and $\Gamma_{-}(\nu,c^{-})$ with $\nu = 16\sigma_{X}^{4}(1+\sqrt{(n-1)/8})^{2}/(n^{2}-n)$, $c^{+} = 6\sigma^{2}_{X}/n$ and $c^{-} = 13\sigma^{2}_{X}/3n$.
\end{proof}

\section{Proof of Section~3}
This section gives the proof of Theorem~\ref{thm:pinf} and Theorem~\ref{thm:pfinite} in Section~3.

\subsection{Proof of Theorem~\ref{thm:pinf}}
\begin{proof}
    Since $\delta = T^{-1}$, we have $\mathbb{P}(\xi^{c}_{\tau}(\delta)) = 2KT^{-1}$, then:
    \begin{align*}
        \mathbb{E}\left[R_{\infty}(\bm{n}_{\pi_1}) - R_{\infty}(\bm{n}^{*})\right] &= 
        \mathbb{E}\left[R_{\infty}(\bm{\lambda}) - R_{\infty}(\bm{\lambda}^{*})\right] \\
        &=  \mathbb{P}(\xi_{\tau}(\delta))\mathbb{E}\left[R_{\infty}(\bm{\lambda}) - R_{\infty}(\bm{\lambda}^{*})|\xi_{\tau}(\delta)\right]+ \mathbb{P}(\xi^{c}_{\tau}(\delta))\mathbb{E}\left[R_{\infty}(\bm{\lambda}) - R_{\infty}(\bm{\lambda}^{*})|\xi^{c}_{\tau}(\delta)\right] \\
        &\leq \mathbb{E}\left[R_{\infty}(\bm{\lambda}) - R_{\infty}(\bm{\lambda}^{*})|\xi_{\tau}(\delta)\right]+ 2KT^{-1}\mathbb{E}\left[R_{\infty}(\bm{\lambda}) - R_{\infty}(\bm{\lambda}^{*})|\xi^{c}_{\tau}(\delta)\right] \\
        &\leq  \frac{1}{T}\left[(K-1)\varepsilon^+_{\tau}+(\frac{\Sigma_{2}}{\sigma^{2}_{\min}}-1)\varepsilon^{-}_{\tau} \right]+o(T^{-3/2})\\
        &\leq \frac{4\sqrt{2}\sigma^{2}(K+\Sigma_{2}/\sigma^{2}_{\min}-2)}{T}\sqrt{\frac{\log T}{\tau}}+o(T^{-3/2}) \\
        &\leq 4\sqrt{2}\sigma^{2}\lambda^{-1/2}(K+\Sigma_{2}/\underline{\sigma}^{2}-2)T^{-3/2}\sqrt{\log T}+o(T^{-3/2})
    \end{align*} 
    where the second inequality is based on Lemma~\ref{lemma:Rgapinf}.
\end{proof}

\subsection{Proof of Theorem~\ref{thm:pfinite}}
\begin{proof}
    Since $\delta = T^{-3/2}$, we have $\mathbb{P}(\xi^{c}_{\tau}(\delta)) = 2KT^{-3/2}$, then:
    \begin{align*}
        \mathbb{E}\left[R_{p}(\bm{n}_{\pi_1}) - R_{p}(\bm{n}^{*})\right] &= 
        \mathbb{E}\left[R_{p}(\bm{\lambda}) - R_{p}(\bm{\lambda}^{*})\right] \\
        &=  \mathbb{P}(\xi_{\tau}(\delta))\mathbb{E}\left[R_{p}(\bm{\lambda}) - R_{p}(\bm{\lambda}^{*})|\xi_{\tau}(\delta)\right]+ \mathbb{P}(\xi^{c}_{\tau}(\delta))\mathbb{E}\left[R_{p}(\bm{\lambda}) - R_{p}(\bm{\lambda}^{*})|\xi^{c}_{\tau}(\delta)\right] \\
        &\leq \mathbb{E}\left[R_{p}(\bm{\lambda}) - R_{p}(\bm{\lambda}^{*})|\xi_{\tau}(\delta)\right]+ 2KT^{-3/2}\mathbb{E}\left[R_{p}(\bm{\lambda}) - R_{p}(\bm{\lambda}^{*})|\xi^{c}_{\tau}(\delta)\right] \\
        &\leq \frac{p^{2}(\Sigma_{q})^{1/p}\Sigma_{q-4}}{2(p+1)T}(\varepsilon^{+}_{\tau})^{2} +o(T^{-2})\\
        &=\frac{24\sigma^{4}p^{2}(\Sigma_{q})^{1/p}\Sigma_{q-4}}{\lambda(p+1)}T^{-2}\log T+o(T^{-2})
    \end{align*} 
    where the second inequality is based on Lemma~\ref{lemma:Rgapfinite}.    
\end{proof}

\section{Proof of Section~4}
This section gives the proof of Theorem~\ref{thm:regretAlg3_subgaussian} in Section~4.

\subsection{Proof of Theorem~\ref{thm:regretAlg3_subgaussian}}
\begin{proof}
For the case $p=\infty$, assuming $\sigma^{2}_{\min} = \sigma^{2}_{K}$, then we have:
\begin{align*}
    \max_{k}\frac{\sigma_k^2}{\lambda_{k,\pi_2}} - \Sigma_2
    &\le \max_{k}\sum_{j\ne k}\big(\sigma_j^2+\varepsilon^+_{\tau_{j}}\big)\big(1+\sum_{n=1}^{\infty}\Big(\frac{\varepsilon^-_{\tau_{k}}}{\sigma_k^2}\Big)^{n}\big)-(\Sigma_2-\sigma_k^2)\\
    &=\sum_{j\neq K}\varepsilon^+_{\tau_{j}}+(\frac{\Sigma_{2}}{\sigma^{2}_{\min}}-1)\varepsilon^{-}_{\tau_{K}}+\mathcal{O}(T^{-1}\log T)\\
    &\leq 8\sigma^{2}\sum_{j\neq K}\sqrt{\frac{\log T}{\alpha_{j}T}} + 8(\frac{\Sigma_{2}}{\sigma^{2}_{\min}}-1)\sigma^{2}\sqrt{\frac{\log T}{\alpha_{K}T}}+\mathcal{O}(T^{-1}\log T) \\
    &= 8\sigma^{2}\left[\sum_{j\neq K}\alpha^{-1/2}_{j}+(\frac{\Sigma_{2}}{\sigma^{2}_{\min}}-1)\alpha^{-1/2}_{K}\right]\sqrt{\frac{\log T}{T}}+\mathcal{O}(T^{-1}\log T) \\
    &\leq  8\sigma^{2}\sqrt{T^{-1}\log T}\left[\sqrt{\Sigma_{2}}\Sigma_{-1}+(\Sigma_{2}\sigma^{-2}_{\min}-2)\sqrt{\Sigma_{2}}\sigma^{-1}_{\min}\right]+\mathcal{O}(T^{-1}\log T)
\end{align*}
where the second inequality is from $\alpha_{k}T \leq \tau_{k}$, the last inequality is based on Lemma~\ref{lemma:general_alpha} and $[1-\Theta(\sqrt{T^{-1}\log T})]^{-1/2} = 1+ \Theta\bigl(\sqrt{T^{-1}\log T}\bigr)$.
Then we can get that:
\begin{align*}
    \mathbb{E}\left[R_{\infty}(\bm{n}_{\pi_2}) - R_{\infty}(\bm{n}^{*})\right] &= 
    \mathbb{E}\left[R_{\infty}(\bm{\lambda}) - R_{\infty}(\bm{\lambda}^{*})\right] \\
    &=  \mathbb{P}(\xi_{T}(\delta))\mathbb{E}\left[R_{\infty}(\bm{\lambda}) - R_{\infty}(\bm{\lambda}^{*})|\xi_{T}(\delta)\right]+ \mathbb{P}(\xi^{c}(\delta))\mathbb{E}\left[R_{\infty}(\bm{\lambda}) - R_{\infty}(\bm{\lambda}^{*})|\xi^{c}_{T}(\delta)\right] \\
    &\leq \mathbb{E}\left[R_{\infty}(\bm{\lambda}) - R_{\infty}(\bm{\lambda}^{*})|\xi_{T}(\delta)\right]+ 2T^{-1}\mathbb{E}\left[R_{\infty}(\bm\lambda) - R_{\infty}(\bm{\lambda}^{*})|\xi^{c}_{T}(\delta)\right] \\
    &\leq 8\sigma^{2}\sqrt{T^{-3}\log T}(\sqrt{\Sigma_{2}}\Sigma_{-1}+(\Sigma_{2}\sigma^{-2}_{\min}-2)\sqrt{\Sigma_{2}}\sigma^{-1}_{\min})+o(T^{-3/2})\\
    &= 8\sigma^{2}\FalgInf{2}{\bm{\sigma^{2}}}\cdot T^{-3/2}\sqrt{\log T}+o(T^{-3/2})     
\end{align*}
When $p$ is finite, then based on the proof of Lemma~\ref{lemma:Rgapfinite}, when $\xi_{T}(\delta)$ exists, we have:
\begin{align*}
    \sum_{k=1}^{K}\frac{(\lambda_{k,\pi_2}-\lambda^{*}_{k})^{2}}{\lambda^{*}_{k}} &\leq \frac{\Sigma_{q}}{(\sum_{k=1}^{K}(\sigma^{2}_{k}-\varepsilon^{-}_{\tau_{k}})^{q/2})^{2}}\sum_{k=1}^{K}\frac{[(\sigma^{2}_{k}+\varepsilon^{+}_{\tau_{k}})^{q/2}-\sigma^{q}_{k}]^{2}}{\sigma_{k}^{q}} \\
    &\leq \frac{q^{2}\sum_{k=1}^{K}\sigma_{k}^{q-4}(\varepsilon^{+}_{\tau_{k}})^{2}}{4\Sigma_{q}}+\mathcal{O}(T^{-3/2}\log^{3/2}T) \\
    &\leq \frac{20\sigma^{4}q^{2}\log T}{\Sigma_{q}T}\sum_{k=1}^{K}\frac{\sigma_{k}^{q-4}}{\alpha_{k}}+\mathcal{O}(T^{-3/2}\log^{3/2}T)\\
    &\leq \frac{20\sigma^{4}q^{2}\log T}{\Sigma_{q}T}\sum_{k=1}^{K}\frac{\sigma_{k}^{q-4}}{\lambda^{*}_{k}}+\mathcal{O}(T^{-3/2}\log^{3/2}T) \\
    &= 20\sigma^{4}q^{2}(\Sigma_{-4})T^{-1}\log T+\mathcal{O}(T^{-3/2}\log^{3/2}T)
\end{align*}
Then we have:
\begin{align*}
    \mathbb{E}\left[R_{\infty}(\bm{n}_{\pi_2}) - R_{\infty}(\bm{n}^{*})\right] &= 
    \mathbb{E}\left[R_{\infty}(\bm{\lambda}) - R_{\infty}(\bm{\lambda}^{*})\right] \\
    &=  \mathbb{P}(\xi_{T}(\delta))\mathbb{E}\left[R_{\infty}(\bm{\lambda}) - R_{\infty}(\bm{\lambda}^{*})|\xi_{T}(\delta)\right]+ \mathbb{P}(\xi^{c}_{T}(\delta))\mathbb{E}\left[R_{\infty}(\bm{\lambda}) - R_{\infty}(\bm{\lambda}^{*})|\xi^{c}_{T}(\delta)\right] \\
    &\leq \mathbb{E}\left[R_{\infty}(\bm{\lambda}) - R_{\infty}(\bm{\lambda}^{*})|\xi_{T}(\delta)\right]+ 2T^{-3/2}\mathbb{E}\left[R_{\infty}(\bm{\lambda}) - R_{\infty}(\bm{\lambda}^{*})|\xi^{c}_{T}(\delta)\right] \\
    &\leq  \frac{40\sigma^{4}p^{2}(\Sigma_{q})^{2/q}(\Sigma_{-4})}{p+1}T^{-2}\log T+o(T^{-2}) \\
    &= 40\sigma^{4}\Falgp{2}{\bm{\sigma}^{2}}\cdot T^{-2}\log T + o(T^{-2})
\end{align*}
\end{proof}

\section{Proof of Section~5}
This section gives the proof of Lemma~\ref{lemma:condition_context_MSE}, Lemma~\ref{lemma:MSE_bound} and Theorem~\ref{thm:Contextual_regret} in Section~5.
\subsection{Proof of Lemma~\ref{lemma:condition_context_MSE}}
\begin{proof}
\begin{align*}
    \hat{\beta}_{k,t} - \beta_{k}&= \big(\gamma I_d+\sum_{s=1}^{t}c_{k,s}c_{k,s}^\top\big)^{-1}\sum_{s=1}^{t}c_{k,s}X_{k,s}-\beta_{k}\\
    &= \big(\gamma I_d+\sum_{s=1}^{t}c_{k,s}c_{k,s}^\top\big)^{-1}\sum_{s=1}^{t}c_{k,s}(\beta_k^\top c_{k,s} + \eta_{k,s}) -\beta_{k}\\
    &= V_{k,t}^{-1}\sum_{s=1}^{t}\eta_{k,s}c_{k,s}-\gamma V_{k,t}^{-1}\beta_{k}.
\end{align*}
Let $\bm{\eta_{k,t}} = (\eta_{k,1},\cdots,\eta_{k,t})^{\top} \in \mathbb{R}^{t}$, and $\bm{C_{k,t}}=[c_{k,1},\cdots,c_{k,t}] \in \mathbb{R}^{d \times t}$, then conditioned on $\bm{C_{k,t}}$, we have:
\begin{align*}
\mathbb{E}\left[\|\hat{\beta}_{k,t} - \beta_{k}\|^{2} \mid \bm{C_{k,t}} \right] 
    &= \mathbb{E}\left[\left\| V_{k,t}^{-1}\sum_{s=1}^{t}\eta_{k,s}c_{k,s}-\gamma V_{k,t}^{-1}\beta_{k} \right\|^{2} \Bigg| \bm{C_{k,t}} \right] \\
    &= \mathbb{E}\left[\gamma^{2}\beta_{k}^{\top}V_{k,t}^{-2}\beta_{k}+\bm{\eta_{k,t}}^{\top}\bm{C_{k,t}}^{\top}V_{k,t}^{-2}\bm{C_{k,t}}\bm{\eta_{k,t}}-2\bm{\eta_{k,t}}^{\top}\bm{C_{k,t}}^{\top}V_{k,t}^{-2}\gamma\beta_{k}\Bigg| \bm{C_{k,t}} \right] \\
    &= \mathbb{E}\left[\gamma^{2}\beta_{k}^{\top}V_{k,t}^{-2}\beta_{k}+\bm{\eta_{k,t}}^{\top}\bm{C_{k,t}}^{\top}V_{k,t}^{-2}\bm{C_{k,t}}\bm{\eta_{k,t}}\Bigg| \bm{C_{k,t}} \right] \\
    &= \gamma^{2}\beta_{k}^{\top}V_{k,t}^{-2}\beta_{k}+\sigma_{k}^{2}\operatorname{Tr}(V_{k,t}^{-2}\bm{C_{k,t}}\bm{C_{k,t}}^{\top}) \\
    &= \gamma^{2}\beta_{k}^{\top}V_{k,t}^{-2}\beta_{k}+\sigma_{k}^{2}\operatorname{Tr}(V_{k,t}^{-2}(V_{k,t}-\gamma I_d)) \\
    &= \gamma^{2}\beta_{k}^{\top}V_{k,t}^{-2}\beta_{k}+\sigma_{k}^{2}\big(\operatorname{Tr}(V_{k,t}^{-1})-\gamma\operatorname{Tr}(V_{k,t}^{-2})\big) \\
    &= \sigma_{k}^{2}\operatorname{Tr}(V_{k,t}^{-1})+   \gamma^{2}\beta_{k}^{\top}V_{k,t}^{-2}\beta_{k}-\gamma\sigma_{k}^{2}\operatorname{Tr}(V_{k,t}^{-2}).
\end{align*}   
\end{proof}

\subsection{Proof of Lemma~\ref{lemma:MSE_bound}}
\begin{proof}
    First let $V_{k,n}(0)=\sum_{s=1}^{n}c_{k,s}c_{k,s}^\top$ when $\gamma=0$. As we set $\gamma = \lambda_{\min}^{\mathcal{C}}/n$, we note that for $V_{k,n}$ we can decide whether to add $\gamma I_d$ or not based on $\sum_{s=1}^{n}c_{k,s}c_{k,s}^\top$ as:
    \begin{align*}
    V_{k,n} = \left[\gamma-\lambda_{\min}(V_{k,n}(0))\right]\mathbb{I}_{\{\lambda_{\min}(V_{k,n)}(0))<\gamma\}}I_d+V_{k,n}(0).
    \end{align*}
    This means that only when $\lambda_{\min}(V_{k,n}(0))$ is very small, we need to add this term. Let $\zeta_{k,n}=\{\lambda_{\min} \left( V_{k,n}(0) \right) \geq \gamma \}$, then based on Lemma~\ref{lemma:condition_context_MSE}, we have:
    \begin{align} \label{beta_MSE}
    \mathbb{E}\left[\|\hat{\beta}_{k,n} - \beta_k\|^{2}\right] &\leq \mathbb{P}\left( \zeta_{k,n}^c \right)
    \cdot \bigg[
        \gamma^{2} \|\beta_k\|^2 \|V_{k,n}^{-2}\|
        + \sigma_k^{2} \cdot \mathbb{E}\left[
            \operatorname{Tr} (V_{k,n}^{-1}) \Big|\, \zeta_{k,n}^c
        \right]
    \bigg] \notag \\
    &\quad + \mathbb{P}\left( \zeta_{k,n} \right)
    \cdot \sigma_k^{2} \cdot \mathbb{E} \left[
        \operatorname{Tr} \left( 
            (V_{k,n}(0))^{-1}
        \right) \Big|\, \zeta_{k,n} 
    \right] \notag\\
    &\leq \mathbb{P}\left( \zeta_{k,n}^c \right)
    \cdot \left[
        \|\beta_k\|^2 + \sigma_k^2 \cdot \frac{d}{\gamma}
    \right]
    + \sigma_k^2 \cdot \mathbb{E} \left[
        \operatorname{Tr} \left( 
            (V_{k,n}(0))^{-1}
        \right) \Big|\, \zeta_{k,n}
    \right].
    \end{align}
    where the second inequality is because $\beta^{\top}A\beta \leq \|A\|\|\beta\|^{2}$ and $\|A^{-2}\| = \|A^{-1}\|^{2} \leq \lambda^{-2}_{\min}(A)$ for any positive definite matrix $A$.

    Then we need to bound $\lambda_{\min}(V_{k,n}(0))$ by using the matrix concentration inequality as Theorem~\ref{thm:matrix_bern}.
    Let $X_s=c_sc_s^{\top}-\Sigma$, then $V_t(0)=t\Sigma+\sum_{s=1}^{t}X_s$. The matrices $\{X_s\}_{s=1}^{t}$ are independent, centered, self–adjoint, and satisfy $\|X_s\|\le\|c_s\|^{2}+\|\Sigma\|\le R^{2}+R^{2}=2R^{2}$. And we have:
    \begin{align*}
        \bigl\| \mathbb{E}[X_s^{2}]\bigr\|
        &= \bigl\| \mathbb{E}[(c_{s}c_{s}^{\top})^{2}]-\Sigma^{2} \bigr\| \\
        &\leq \bigl\| \mathbb{E}[(c_{s}c_{s}^{\top})^{2}]\bigr\| +\bigl\|  \Sigma^{2} \bigr\| \\
        &\leq R^{2}\bigl\|\Sigma\bigr\| + \bigl\|  \Sigma^{2} \bigr\| \\
        &\leq 2R^{2}\bigl\|\Sigma\bigr\| 
    \end{align*}
    where the first equality is due to $ \mathbb{E}[c_{s}c_{s}^{\top}\Sigma] = \Sigma^{2}$, the second inequality is because $\mathbb{E}[(c_{s}c_{s}^{\top})^{2}] \preceq R^{2}\mathbb{E}[c_{s}c_{s}^{\top}]$ based on Assumption~\ref{assump:postive_definite}, the third inequality is from $\|  \Sigma^{2} \| = \|  \Sigma \|^{2} \leq R^{2}\|  \Sigma \|$. Then we have $v = \| \sum_{s=1}^{t}\mathbb{E}[X_s^{2}]\| \leq 2tR^{2}\|\Sigma\|$. Let $S_{t} = \sum_{s=1}^{t}X_s$, and assume $m \geq 2$, by Theorem~\ref{thm:matrix_bern} we have:    
    \begin{align*}
        \mathbb{P}(\{\|S_{t}\|\ge \tfrac{m-1}{m} t\lambda_{\min}(\Sigma)\}) 
        &\leq 2d \cdot \exp(-\frac{\tfrac{(m-1)^{2}}{2m^{2}} t^{2}\lambda^{2}_{\min}(\Sigma)}{2tR^{2}\|\Sigma\|+\tfrac{2(m-1)}{3m} R^{2}t\lambda_{\min}(\Sigma)}) \\
        &\leq  2d \cdot \exp(-\frac{3(m-1)^{2}t\lambda^{2}_{\min}(\Sigma)}{4m(4m-1)R^{2}\|\Sigma\|}) \\
        &\leq 2d \cdot \exp(-t\kappa(m)) 
    \end{align*}
   Then based on the fact that $\lambda_{\min}(V_{k,n}(0)) = \lambda_{\min}(n\Sigma+S_{n}) \geq n\lambda_{\min}(\Sigma) - \|S_{n}\|$, we have:
    \begin{align*}
        \mathbb{P}(\zeta_{n}^{c}) &\leq  \mathbb{P}(\{\|S_{n}\|\ge (1-n^{-2}) n\lambda_{\min}(\Sigma)\})\\
        &\leq 2d\exp(-n\cdot \kappa(n^{2})) \\
        &\leq 2d \exp(-\frac{3n\lambda^{2}_{\min}(\Sigma)}{56R^{2}\|\Sigma\|})
    \end{align*}
    where the last inequality is based on $n \geq 2$. When $\zeta_{n}$ exists, let $u_{0} = 2/(n\lambda_{\min}(\Sigma))$, we can get:
    \begin{align*}
        \mathbb{E} \left[\operatorname{Tr} \left( V_n(0)^{-1}\right)\Big|\, \zeta_n\right] &\leq \mathbb{E}\left[\frac{d}{\lambda_{\min}(V_{t}(0))}\Big|\, \zeta_t\right] \\
        &= d\int_{0}^{u_{0}} \mathbb{P}(\lambda_{\min}(V_{n}(0))\leq \frac{1}{u}) du + d\int_{u_0}^{\tfrac{n}{\lambda_{\min}(\Sigma)}} \mathbb{P}(\lambda_{\min}(V_{n}(0))\leq \frac{1}{u}) du \\        
        &\leq \frac{2d}{n\lambda_{\min}(\Sigma)} + \frac{d}{n\lambda_{\min}(\Sigma)}\int_{2}^{n^{2}} \mathbb{P}(\lambda_{\min}(V_{n}(0))\leq \frac{n}{m}\lambda_{\min}(\Sigma))dm \\
        &\leq \frac{2d}{n\lambda_{\min}(\Sigma)} + \frac{d}{n\lambda_{\min}(\Sigma)}\int_{2}^{n^{2}} \mathbb{P}(\|S_{n}\| \geq \tfrac{m-1}{m}n\lambda_{\min}(\Sigma))dm \\
        &\leq \frac{2d}{n\lambda_{\min}(\Sigma)} + \frac{2d^{2}}{n\lambda_{\min}(\Sigma)}\int_{2}^{n^{2}} \exp(-\kappa(m)n)dm \\
        &\leq \frac{2d}{n\lambda_{\min}(\Sigma)}+\frac{2d^{2}(n^{2}-2)}{n\lambda_{\min}(\Sigma)}\exp(-\frac{3n\lambda^{2}_{\min}(\Sigma)}{56R^{2}\|\Sigma\|}).
    \end{align*}
    Then combine them into equation~\eqref{beta_MSE}, and let $\lambda^{\mathcal{C}}_{\min} = \lambda_{\min}(\Sigma)$, we have:
    \begin{align*}
        \mathbb{E}\left[\|\hat{\beta}_{k,n} - \beta_k\|^{2}\right] &\leq  2d \cdot \exp(-\frac{3n(\lambda_{\min}^{\mathcal{C}})^{2}}{56R^{2}\|\Sigma\|})\cdot \left[\|\beta_k\|^2 + \sigma_k^2 \cdot \frac{dn}{\lambda^{\mathcal{C}}_{\min}}\right] \\
        &\quad + \sigma_k^2 \cdot \left[ \frac{2d}{n\lambda^{\mathcal{C}}_{\min}}+\frac{2d^{2}n}{\lambda^{\mathcal{C}}_{\min}}\exp(-\frac{3n(\lambda^{\mathcal{C}}_{\min})^{2}}{56R^{2}\|\Sigma\|})\right] \\
        &\leq \frac{2d\sigma_{k}^{2}}{n\lambda^{\mathcal{C}}_{\min}} + \exp(-\frac{3n(\lambda^{\mathcal{C}}_{\min})^{2}}{56R^{2}\|\Sigma\|}) \cdot \left[2d\|\beta_k\|^2 + \frac{4n d\sigma_k^2}{\lambda^{\mathcal{C}}_{\min}} \right]\\
        &= \frac{2d\sigma_{k}^{2}}{n\lambda^{\mathcal{C}}_{\min}} + o(n^{-2}_{k})
    \end{align*}
\end{proof}

\subsection{Proof of Theorem~\ref{thm:Contextual_regret}}
\begin{proof}
   This result is directly from Lemma~\ref{lemma:MSE_bound} and Theorem~\ref{thm:regretAlg3_subgaussian}. 
\end{proof}

\section{Proof of Appendix}
This section gives the proof of lemmas and theorems shown in Appendix, which includes Lemma~\ref{lemma:subgamma}, Lemma~\ref{lemma:Rgapinf}, Lemma~\ref{lemma:Rgapfinite}, Lemma~\ref{lemma:general_alpha}, 
Theorem~\ref{thm:strictly_p} and Theorem~\ref{thm:regretAlg3_strictlysubgaussian}.

\subsection{Proof of Lemma~\ref{lemma:subgamma}}
\begin{proof}
For the general $\sigma$-subgaussian case, for the right tail with $t>0$, we have:
\begin{align*}
    \mathbb{E}[\exp(t(X^{2}-\mathbb{E}[X^{2}]))] &=\exp(-t\sigma_{X}^{2})\big(1+t\mathbb{E}[(X^{2})]+\sum_{i=2}^{\infty}\frac{t^{i}\mathbb{E}[X^{2i}]}{i !}\big)\\
    &\leq \exp(-t\sigma_{X}^{2})\big(1+t\sigma_{X}^{2}+ \sum_{i=2}^{\infty}t^{i} 2^{i+1}\sigma_{\mathrm{s}}^{2i}\big)\\
    &= \exp(-t\sigma_{X}^{2})\big(1+t\sigma_{X}^{2}+ \frac{8\sigma^{4}t^{2}}{1-2\sigma_{\mathrm{s}}^{2}t} \big)\\
    &\leq \exp(\frac{8\sigma^{4}t^{2}}{1-2\sigma_{\mathrm{s}}^{2}t})
\end{align*}
where the first inequality is based on $\mathbb{E}[X^{2q}] \leq 2q!2^{q}\sigma^{2q}$ for integer $q\geq 1$, which is Theorem 2.1 in \cite{boucheron2013concentration}, the last inequality is from $1+x \leq e^{x}$. For the left tail, with $t>0$ we have:
\begin{align*}
    \mathbb{E}[\exp(t(\mathbb{E}[X^{2}]-X^{2}))] &\leq 1+t\mathbb{E}[(\mathbb{E}[X^{2}]-X^{2})]+\frac{\exp(t\sigma_{X}^{2})-t\sigma_{X}^{2}-1}{\sigma_{X}^{4}}\mathbb{V}(X^{2})
    \\
    &= 1+\frac{\exp(t\sigma_{X}^{2})-t\sigma_{X}^{2}-1}{\sigma_{X}^{4}}\mathbb{V}(X^{2})
\end{align*}
where the first inequality is from Theorem~2.9 in \cite{boucheron2013concentration}. For $0<t<3/\sigma^{2}\leq 3/\sigma_{X}^{2}$, we have:
\begin{align*}
    \psi_{\mathbb{E}[X^{2}]-X^{2}}(t) &\leq \frac{\exp(t\sigma_{X}^{2})-t\sigma_{X}^{2}-1}{\sigma_{X}^{4}}\mathbb{V}(X^{2}) \\
    &\leq \frac{\mathbb{V}(X^{2})}{\sigma_{X}^{4}} \frac{\sigma_{X}^{4}t^{2}}{2(1-t\sigma_{X}^{2}/3)} \\
    &\leq \frac{8\sigma^{4}t^{2}}{1-t\sigma^{2}/3}
\end{align*}
where the first inequality is from $\log(1+x)\leq x$, the second inequality is from $e^{x}-x-1\leq x^{2}/(2-2x/3)$ for $0<x<3$, the last inequality is from $\mathbb{V}(X^{2}) \leq \mathbb{E}[X^{4}] \leq 16\sigma^{4}$.

For the special case $\sigma_{X}^{2}=\sigma^{2}$, for the right tail with $t>0$, we have:
\begin{align*}
\mathbb{E}[\exp(t(X^{2}-\mathbb{E}[X^{2}]))] 
&= \frac{\exp(-t\sigma_{X}^{2})}{\sqrt{4\pi t}} \int_{-\infty}^{\infty} \mathbb{E}[\exp(sX)] \exp\left(-\frac{s^{2}}{4t}\right) ds \\
&\leq \frac{\exp(-t\sigma_{X}^{2})}{\sqrt{4\pi t}} \int_{-\infty}^{\infty} \exp\left( \frac{\sigma_{X}^{2}s^{2}}{2} - \frac{s^{2}}{4t} \right) ds \\
&= \frac{\exp(-t\sigma_{X}^{2})}{\sqrt{4\pi t}} \cdot \sqrt{ \frac{4\pi t}{1 - 2\sigma_{X}^{2} t} } \\
&\leq \exp\left(\frac{\sigma_{X}^{4}t^{2}}{1-2\sigma_{X}^{2}t}\right)
\end{align*}
where the first equality is from Hubbard–Stratonovich transformation, the first inequality uses the subgaussian property, the second equality is based on \( \int_{-\infty}^{\infty} \exp(-a s^2) ds = \sqrt{\pi / a} \) for \( a > 0 \), the last inequality is by $-\log(1-u)-u \leq u^{2}/[2(1-u)]$ for $0<u<1$. For the left tail, the proof is similar as the general case, only in the last step we have $\mathbb{V}(X^{2}) = \mathbb{E}[X^{4}]-\sigma_{X}^{4}\leq2\sigma_{X}^{4}$.
\end{proof}

\subsection{Proof of Lemma~\ref{lemma:Rgapinf}}
\begin{proof}
    If $\xi_{\tau}(\delta)$ holds and $\varepsilon^{-}_{\tau}(\delta) < \sigma^{2}_{\min}$, then we have:
\begin{align*}
    R_{\infty}(\bm{\lambda}_{\pi_1}) - R_{\infty}(\bm{\lambda}^{*}) &= 
    \frac{1}{T}\left(\max_{k}\frac{\sigma_k^2}{\lambda_k} - \Sigma_{2}\right) \\
    &= \frac{1}{T}\left(\max_{k}\frac{\sigma_k^2\,\hat{\Sigma}_{2,\tau}}{\hat{\sigma}_{k,\tau}^2} - \Sigma_2\right) \\
    &\le \frac{1}{T}\left(\max_{k}\frac{\Sigma_{2}-\sigma^{2}_{k}+(K-1)\varepsilon^{+}_{\tau}
    }{\,1-\varepsilon^-_{\tau}/\sigma_{k}^2\,} - (\Sigma_2-\sigma_k^2)\right)\\
    &= \frac{1}{T}\left(\frac{\Sigma_{2}-\sigma^{2}_{\min}+(K-1)\varepsilon^{+}_{\tau}
    }{\,1-\varepsilon^-_{\tau}/\sigma_{\min}^2\,} - (\Sigma_2-\sigma_{\min}^2)\right)\\    
    &= \frac{1}{T}\left[(K-1)\varepsilon^+_{\tau}+\frac{\varepsilon^{-}_{\tau}}{\sigma^{2}_{\min}}(\Sigma_{2}-\sigma^{2}_{\min})
    +(K-1)\varepsilon^+_{\tau} \cdot \sum_{n=1}^{\infty}\Big(\frac{\varepsilon^-_{\tau}}{\sigma_{\min}^2}\Big)^{n}+(\Sigma_{2}-\sigma^{2}_{\min})\sum_{n=2}^{\infty}\Big(\frac{\varepsilon^-_{\tau}}{\sigma_{\min}^2}\Big)^{n}\right]\\
    &=\frac{1}{T}\left[(K-1)\varepsilon^+_{\tau}+(\frac{\Sigma_{2}}{\sigma^{2}_{\min}}-1)\varepsilon^{-}_{\tau}+\frac{(K-1)\varepsilon^+_{\tau}\varepsilon^-_{\tau}}{\sigma^{2}_{\min}-\varepsilon^-_{\tau}}+\frac{(\Sigma_{2}-\sigma^{2}_{\min})(\varepsilon^{-}_{\tau})^{2}}{\sigma^{2}_{\min}(\sigma^{2}_{\min}-\varepsilon^{-}_{\tau})}\right]\\
    &= \frac{1}{T}\left[ (K-1)\varepsilon^+_{\tau}+(\frac{\Sigma_{2}}{\sigma^{2}_{\min}}-1)\varepsilon^{-}_{\tau}\right]+o(T^{-3/2})
\end{align*}
where for the third equality, the maximization is attained when we choose $\hat{\sigma}^{2}_{K,\tau} = \sigma^{2}_{K}-\varepsilon^{-}_{\tau}(\delta)$ for the lowest--variance arm $K$ and $\hat{\sigma}^{2}_{k,\tau} = \sigma_{k}^{2}+\varepsilon^{+}_{\tau}(\delta)$ for the other arms, the fourth equality is from $(1-x)^{-1} = \sum_{n=0}^{\infty} x^{n}$ for $x<1$.
\end{proof}

\subsection{Proof of Lemma~\ref{lemma:Rgapfinite}}
\begin{proof}
    Since $\varepsilon^{-}_{\tau} < \sigma^{2}_{\min}$, assume $\varepsilon^{-}_{\tau} \leq \sigma^{2}_{\min}/m$ for $m>1$. Based on Lemma~\ref{lemma:Rgaptaylor}, we first bound the term 
    $\sum_{k=1}^{K} \frac{(\lambda_{k,\pi_1} - \lambda_k^*)^2}{\lambda_k^*}$. Then, we control the third-order remainder of Taylor expansion. Firstly, as we have $\sigma^{2}_{\min}$
    by using a slightly loose upper bound, we have:
    \begin{align*}
        \sum_{k=1}^{K}\frac{(\lambda_{k,\pi_1}-\lambda^{*}_{k})^{2}}{\lambda^{*}_{k}}
        &= \sum_{k=1}^{K} \frac{[(\hat{\sigma}_{k,\tau}^{q}-\sigma_{k}^{q})-\lambda_{k}^{*}(\hat{\Sigma}_{q,\tau}-\Sigma_{q})]^{2}}{\lambda^{*}_{k}(\hat{\Sigma}_{q,\tau})^{2}} \\
        &\leq \frac{\Sigma_{q}}{(\hat{\Sigma}_{q,\tau})^{2}}\sum_{k=1}^{K}\frac{(\hat{\sigma}^{q}_{k,\tau}-\sigma_{k}^{q})^{2}}{\sigma_{k}^{q}} \\
        &\leq \frac{\Sigma_{q}}{(\sum_{k=1}^{K}(\sigma^{2}_{k}-\varepsilon^{-}_{\tau})^{q/2})^{2}}\sum_{k=1}^{K}\frac{[(\sigma^{2}_{k}+\varepsilon^{+}_{\tau})^{q/2}-\sigma^{q}_{k}]^{2}}{\sigma_{k}^{q}} \\
        &\leq \frac{q^{2}(\varepsilon^{+}_{\tau})^{2}\Sigma_{q-4}}{4\Sigma_{q}}\sum_{n=0}^{\infty}(n+1)(\tfrac{mq\varepsilon^{-}_{\tau}\Sigma_{q-2}}{2(m-1)\Sigma_{q}})^{n}\\
        &= \frac{q^{2}(\varepsilon^{+}_{\tau})^{2}\Sigma_{q-4}}{4\Sigma_{q}}+o(T^{-1})
\end{align*} 
    where the second inequality is from $(\sigma^{2}_{k}+\varepsilon^{+}_{\tau})^{q/2}-\sigma^{q}_{k} \leq q\sigma^{q-2}_{k}\varepsilon^{+}_{\tau}/2$, the third inequality is from $(\sigma^{2}_{k}-\varepsilon^{-}_{\tau})^{q/2} \geq \sigma^{q}_{k}-mq\varepsilon^{-}_{\tau}\sigma^{q-2}_{k}/(2m-2)$. 
    Then for the reminder term, first for $||\bm{\lambda}_{\pi_1} - \bm{\lambda}^{*}||^3_{\infty}$ we can get that:
    \begin{align*}
    ||\bm{\lambda}_{\pi_1} - \bm{\lambda}^{*}||^3_{\infty} 
    &=  \max_{k}\left|\frac{\hat\sigma_k^q}{\hat\Sigma_q} - \frac{\sigma_k^q}{\Sigma_q}\right|^{3} \\
    &\le \max_{k} \left(\frac{|\hat\sigma_k^q - \sigma_k^q|}{\hat\Sigma_q}
    + \frac{\sigma_k^q |\hat\Sigma_q - \Sigma_q|}{\hat\Sigma_q \, \Sigma_q}\right)^{3}  \\
    &\le \max_{k} \left[\frac{\tfrac{q}{2}(\sigma_k^2 - \varepsilon_{\tau}^-)^{\tfrac q2 - 1}|\sigma^{2}_{k}-\hat\sigma_k^2|}{ \sum_{j=1}^K (\sigma_j^2 - \varepsilon^-_{\tau})^{q/2}}
    + \frac{\sigma_k^q\varepsilon^{+}_{\tau}\sum_{j=1}^K \tfrac{q}{2}(\sigma_j^2 - \varepsilon^-_{\tau})^{\tfrac q2 - 1}}{ \sum_{j=1}^K (\sigma_j^2 - \varepsilon^-_{\tau})^{q/2}\Sigma_q}\right]^{3} \\
    &\le \left[\frac{q(K+1)} {2K\sigma^{2}_{\min}(1-1/m)}\right]^{3}(\varepsilon^{+}_{\tau})^{3} 
    \end{align*}
    where the third inequality is from $(\sigma_j^2 - \varepsilon^-_{\tau})^{\tfrac q2 - 1} \leq 1$ and $|\sigma^{2}_{k}-\hat\sigma_k^2|\leq \varepsilon^{+}_{\tau}$. Then for the term $\max_{k}\frac{\lambda_{k}^{*}}{\lambda_{k,\pi_{1}}}$, we have:
    \begin{align*}
    \max_{k}\frac{\lambda_{k}^{*}}{\lambda_{k,\pi_1}} 
    &= \max_{k} \frac{\sigma_{k}^{q}}{\Sigma_{q}}/(\frac{\hat{\sigma}_{k}^{q}}{\hat{\Sigma}_{q}}) \\
    &\leq \max_{k} \frac{\sigma_{k}^{q}}{\Sigma_{q}}(1+(\sigma^{2}_{k}-\varepsilon^{-}_{\tau})^{-q/2}\sum_{j\neq k }(\sigma^{2}_{j}+\varepsilon^{+}_{\tau})^{q/2})\\
    &\leq \max_{k} \frac{1}{\Sigma_{q}}(\sigma_{k}^{q}+(1-1/m)^{-q/2}\sum_{j\neq k }(\sigma^{2}_{j}+\varepsilon^{+}_{\tau})^{q/2})\\
    &\leq \max_{k} \frac{1}{\Sigma_{q}}(\sigma_{k}^{q}+(1-1/m)^{-q/2}\sum_{j\neq k }(\sigma^{q}_{j}+q\sigma^{q-2}_{j}\varepsilon^{+}_{\tau}/2) \\
    &\leq \max_{k}  \frac{1}{\Sigma_{q}}(\sigma_{k}^{q}+(1-1/m)^{-q/2}(\Sigma_{q}-\sigma_{k}^{q}+q\varepsilon^{+}_{\tau}(\Sigma_{q-2}-\sigma^{q-2}_{k})/2)\\
    &\leq \max_{k}  (1-1/m)^{-q/2}(1+\frac{q(\Sigma_{q-2}-\sigma^{q-2}_{k})}{2\Sigma_{q}}\varepsilon^{+}_{\tau})\\
    &\leq (1-1/m)^{-q/2}(1+\frac{q\Sigma_{q-2}}{2\Sigma_{q}}\varepsilon^{+}_{\tau})\\
    &\leq (1-1/m)^{-q/2}+\mathcal{O}(\sqrt{T^{-1}\log T})
    \end{align*}
    where the third inequality is from $(\sigma^{2}_{k}+\varepsilon^{+}_{\tau})^{q/2}-\sigma^{q}_{k} \leq q\sigma^{q-2}_{k}\varepsilon^{+}_{\tau}/2$. Combine them together, we have:
    \begin{align*}
        R_{p}(\bm{\lambda}_{\pi_1}) - R_{p}(\bm{\lambda}^{*}) &\leq \frac{(p+1)R_{p}(\bm{\lambda}^{*})}{2}\sum_{k=1}^{K}\frac{(\lambda_{k,\pi_1}-\lambda^{*}_{k})^{2}}{\lambda^{*}_{k}} + \frac{7(p+2)^{2}}{\lambda_{\min}^{*}T}\max\limits_{k}(\frac{\lambda_{k}^{*}}{\lambda_{k,\pi_1}})^{3p+3}||\bm{\lambda}_{\pi_1} - \bm{\lambda}^{*}||^3_{\infty} \\
        &\leq  \frac{(p+1)(\Sigma_{q})^{2/q}}{2T}\bigg[\frac{q^{2}(\varepsilon^{+}_{\tau})^{2}\Sigma_{q-4}}{4\Sigma_{q}}\bigg]+o(T^{-2})\\
        &= \frac{p^{2}(\Sigma_{q})^{1/p}\Sigma_{q-4}}{2(p+1)T}(\varepsilon^{+}_{\tau})^{2}+o(T^{-2})
    \end{align*}
\end{proof}

\subsection{Proof of Lemma~\ref{lemma:general_alpha}}
\begin{proof}
Based on $\xi_{T}(\delta)$, we have:
\begin{align*}
   \sigma^{2}_{k} - \varepsilon^{-}_{n}-\varepsilon^{+}_{n} \leq \text{LCB}_{k,n} \leq \sigma^{2}_{k} \leq \text{UCB}_{k,n} \leq \sigma^{2}_{k} + \varepsilon^{-}_{n}+\varepsilon^{+}_{n}.
\end{align*}
Assume $e_{\tau_k} = \varepsilon^{+}_{\tau_k}+\varepsilon^{-}_{\tau_k} \leq \sigma^{2}_{k}/m_{k,\tau_k}$ for $m_{k,\tau_k} >1$, and we can get that:
\begin{align*}
    \tau_{k} &\geq \alpha_{k} T\\
    &\geq \frac{(\sigma^{2}_{k}-e_{\tau_{k}})^{q/2}}{(\sigma^{2}_{k}-e_{\tau_{k}})^{q/2}+\sum_{j\neq k}(\sigma^{2}_{j}+e_{\tau_{j}})^{q/2}}T \\
    &\geq \frac{(\sigma^{2}_{k}-\sigma^{2}_{k}/m_{k,\tau_{k}})^{q/2}}{(\sigma^{2}_{k}-\sigma^{2}_{k}/m_{k,\tau_{k}})^{q/2}+\sum_{j\neq k}(\sigma^{2}_{j}+\sigma^{2}_{j}/m_{j,\tau_{j}})^{q/2}}T \\
    &\geq \min_{j\neq k}\left[\frac{1-1/m_{k,\tau_{k}}}{1+1/m_{j,\tau_j}}\right]^{q/2}\lambda^{*}_{k}T 
\end{align*}
We can get $\tau_{k} = \Omega(T)$, and since $\tau_{k} \leq \lambda^{*}_{k}T$, we have $\tau_{k} = \Theta(T)$.
Then we need to derive the relationship between $m_{k,\tau_k}$ and $T$ for all $k= 1, \cdots, K$:
\begin{align*}
    m_{k,\tau_k}
    &= \frac{\sigma^{2}_{k}}{\varepsilon^{+}_{\tau_k}+\varepsilon^{-}_{\tau_k}}\\
    &= \frac{\sigma^{2}_{k}}{\sigma^{2}}
       \left(
       8\sqrt{2}\,\frac{1+\sqrt{\tau_k-1}}{\sqrt{\tau_k-1}}
       \sqrt{\frac{\alpha_{0}\log T}{\tau_k}}
       +\frac{19\alpha_{0}\log T}{3\tau_k}
       \right)^{-1}\\
    &\ge
    \frac{\sigma^{2}_{k}}{\sigma^{2}}
    \frac{1}{
      8\sqrt{2}\Big(1+\frac{1}{\sqrt{\tau_k-1}}\Big)
      \sqrt{\frac{\alpha_{0}\log T}{\tau_k}}
    }\left[
      1-\frac{ \frac{19\,\alpha_{0}\log T}{3\tau_k} }{
      8\sqrt{2}\Big(1+\frac{1}{\sqrt{\tau_k-1}}\Big)
      \sqrt{\frac{\alpha_{0}\log T}{\tau_k}} }
    \right]\\
    &= \frac{\sigma^{2}_{k}}{\sigma^{2}}
    \frac{1}{8\sqrt{2}}
    \frac{1}{\Big(1+\frac{1}{\sqrt{\tau_k-1}}\Big)}
    \sqrt{\frac{\tau_k}{\alpha_{0}\log T}}
    \left[
      1-\frac{19}{24\sqrt{2}}
      \frac{\sqrt{\alpha_{0}\log T/\tau_k}}{1+\frac{1}{\sqrt{\tau_k-1}}}
    \right]\\
    &= \frac{\sigma^{2}_{k}}{\sigma^{2}}
    \bigg[\frac{1}{8\sqrt{2}}
      \frac{1}{1+\frac{1}{\sqrt{\tau_k-1}}}
      \sqrt{\frac{\tau_k}{\alpha_{0}\log T}}-\frac{19}{384}
      \frac{1}{\Big(1+\frac{1}{\sqrt{\tau_k-1}}\Big)^{2}}
    \bigg]\\
    &\geq \frac{\sigma^{2}_{k}}{\sigma^{2}}
    \bigg[\frac{1}{8\sqrt{2}}
      \frac{1}{1+\frac{1}{\sqrt{\tau_k-1}}}
      \sqrt{\frac{\tau_k}{\alpha_{0}\log T}}-\frac{19}{384}\bigg] \\
    &= \Omega \left(\sqrt{\frac{T}{\log T}}\right)  
\end{align*}
where $\alpha_{0} = 2$ when $p=\infty$ and $\alpha_{0} = 5/2$ when $p<\infty$, the first inequality is from $(A+B)^{-1} \geq A^{-1}(1-B/A)$ for $0\leq B \leq A$. Also, it is easy to verify that $$m_{k,\tau_{k}} \leq \frac{\sigma_{k}^{2}}{\sigma^{2}}\frac{1}{8\sqrt{2}}\sqrt{\frac{\tau_k}{\alpha_{0}\log T}} = \mathcal{O}\left(\sqrt{\frac{T}{\log T}}\right).$$
Then we have $m_{k,\tau_k} = \Theta(\sqrt{\frac{T}{\log T}})$. Let $ \underline{m}_{k} = \min_{j\neq k}m_{j,\tau_j}$, then $\overline{m_{k}} = \Theta(\sqrt{\frac{T}{\log T}})$, then we can get:
\begin{align*}
    \alpha_{k} &=  \min_{j\neq k}\left[\frac{1-1/m_{k,\tau_{k}}}{1+1/m_{j,\tau_j}}\right]^{q/2}\lambda^{*}_{k} \\
    &= \left(\frac{1-1/m_{k,\tau_{k}}}{1+1/\underline{m_{k}}}\right)^{q/2}\lambda^{*}_{k}\\
    &= \lambda^{*}_{k}\left[1-\frac{q}{2}(1/m_{k,\tau_{k}}+1/\underline{m_{k}})+\mathcal{O}(T^{-1}\log T)\right].
\end{align*}
where the third equality is from Taylor expansion of $(1-x)^{a}$ and $(1+x)^{-a}$.
Then assume $$m_{k,\tau_k}, \underline{m_{k}} \in [c_{1}\sqrt{T/ \log T}, c_{2}\sqrt{T/ \log T}],$$ 
then we can get:
\begin{align*}
    \lambda^{*}_{k}\left[1-\frac{q}{c_{2}}\sqrt{T^{-1}\log T}\right] \leq \alpha_{k} \leq \lambda^{*}_{k}\left[1-\frac{q}{c_{1}}\sqrt{T^{-1}\log T} + \mathcal{O}(T^{-1}\log T)\right].
\end{align*}
This shows that $\alpha_{k}= \lambda^{*}_{k}(1-\Theta(\sqrt{T^{-1}\log T}))$.
\end{proof}

\subsection{Proof of Theorem~\ref{thm:strictly_p}}
\begin{proof}
    For $p=\infty$, in the event $\xi_{\tau}(\delta)$, we would have:
    \begin{align*}
    \max_{k}\frac{\sigma_k^2}{\lambda_{k,\pi_1}} - \sum_{j=1}^K \sigma_j^2
    &= \max_{k} \frac{\sigma_k^2\,\hat{\Sigma}_{2,\tau}}{\hat{\sigma}_{k,\tau}^2} - \Sigma_2 \\
    &\le \max_{k} \sum_{j\ne k}\big(\sigma_j^2+\varepsilon^+_{j,\tau}\big)\big(1-\frac{\varepsilon^-_{k,\tau}}{\sigma_k^2}\big)^{-1}-(\Sigma_2-\sigma_k^2)\\
    &= (\frac{1+s^{+}_{\tau}}{1-s^{-}_{\tau}}-1)(\Sigma_{2}-\sigma^{2}_{\min})
    \\
    &\leq (s^{+}_{\tau}+s^{-}_{\tau})(\Sigma_{2}-\sigma^{2}_{\min})+o(T^{-1/2})
    \end{align*}
    Then we have:
        \begin{align*}
        \mathbb{E}\left[R_{\infty}(\bm{n}_{\pi_1}) - R_{\infty}(\bm{n}^{*})\right] &= 
        \mathbb{E}\left[R_{\infty}(\bm{\lambda}) - R_{\infty}(\bm{\lambda}^{*})\right] \\
        &\leq \mathbb{E}\left[R_{\infty}(\bm{\lambda}) - R_{\infty}(\bm{\lambda}^{*})|\xi_{\tau}(\delta)\right]+ 2KT^{-1}\mathbb{E}\left[R_{\infty}(\bm{\lambda}) - R_{\infty}(\bm{\lambda}^{*})|\xi^{c}_{\tau}(\delta)\right] \\
        &\leq  \frac{1}{T}\left[(s^{+}_{\tau}+s^{-}_{\tau})(\Sigma_{2}-\sigma^{2}_{\min})\right]+o(T^{-3/2})\\
        &\leq \frac{4(\Sigma_{2}-\sigma^{2}_{\min})}{T}\sqrt{\frac{\log T}{\tau}}+o(T^{-3/2}) \\
        &\leq 4\lambda^{-1/2}(\Sigma_{2}-\sigma^{2}_{\min})T^{-3/2}\sqrt{\log T}+o(T^{-3/2})
    \end{align*} 
    When $p$ is finite, in the event $\xi_{\tau}(\delta)$, we would have:
    \begin{align*}
        \sum_{k=1}^{K}\frac{(\lambda_{k,\pi_1} -\lambda^{*}_{k})^{2}}{\lambda^{*}_{k}}
        &\leq \frac{\Sigma_{q}}{(\sum_{k=1}^{K}(\sigma^{2}_{k}-\varepsilon^{-}_{k,\tau})^{q/2})^{2}}\sum_{k=1}^{K}\frac{[(\sigma^{2}_{k}+\varepsilon^{+}_{k,\tau})^{q/2}-\sigma^{q}_{k}]^{2}}{\sigma_{k}^{q}} \\
        &= \frac{[(1+s^{+}_{\tau})^{q/2}-1]^{2}}{(1-s^{-}_{\tau})^{q}}\\
        &\leq \frac{q^{2}(s^{+}_{\tau})^{2}}{4(1-s^{-}_{\tau})^{q}}\\
        &\leq \frac{q^{2}(s^{+}_{\tau})^{2}}{4} +o(T^{-1})
    \end{align*}
    Then we have:
    \begin{align*}
        \mathbb{E}\left[R_{p}(\bm{n}_{\pi_1}) - R_{p}(\bm{n}^{*})\right] &= 
        \mathbb{E}\left[R_{p}(\bm{\lambda}) - R_{p}(\bm{\lambda}^{*})\right] \\
        &\leq \mathbb{E}\left[R_{p}(\bm{\lambda}) - R_{p}(\bm{\lambda}^{*})|\xi_{\tau}(\delta)\right]+ 2KT^{-3/2}\mathbb{E}\left[R_{p}(\bm{\lambda}) - R_{p}(\bm{\lambda}^{*})|\xi^{c}_{\tau}(\delta)\right] \\
        &\leq \frac{p^{2}(\Sigma_{q})^{2/q}}{2(p+1)T}(s^{+}_{\tau})^{2} +o(T^{-2})\\
        &=\frac{3p^{2}(\Sigma_{q})^{2/q}}{\lambda(p+1)}T^{-2}\log T+o(T^{-2})
    \end{align*} 
\end{proof}

\subsection{Proof of Theorem~\ref{thm:regretAlg3_strictlysubgaussian}}
\begin{proof}
For $p=\infty$, in the event $\xi_{T}(\delta)$, we would have:
    \begin{align*}
    \max_{k}\frac{\sigma_k^2}{\lambda_{k,\pi_2}} - \Sigma_2
    &\le \max_{k}\sum_{j\ne k}\big(\sigma_j^2+\varepsilon^+_{\tau_{j}}\big)\big(1+\sum_{n=1}^{\infty}\Big(\frac{\varepsilon^-_{\tau_{k}}}{\sigma_k^2}\Big)^{n}\big)-(\Sigma_2-\sigma_k^2)\\
    &= (\Sigma_{2}-\sigma^{2}_{\min})s^{-}_{\tau_{K}}+\sum_{j\neq K}s_{\tau_{j}}^{+}\sigma^{2}_{j}+\mathcal{O}(T^{-1}\log T) \\
    &= 2(\Sigma_{2}-\sigma^{2}_{\min})\sqrt{\frac{\log(1/\delta)}{\tau_{K}}}+ 2\sum_{j\neq K}\sigma^{2}_{j}\sqrt{\frac{\log(1/\delta)}{\tau_{j}}}+\mathcal{O}(T^{-1}\log T) \\
    &\leq \sqrt{8T^{-1}\log T}\frac{\Sigma_{2}-2\sigma^{2}_{\min}}{\sqrt{\alpha_{K}}}+\sqrt{8T^{-1}\log T}\sum_{k=1}^{K} \frac{\sigma^{2}_{k}}{\sqrt{\alpha_{k}}}+\mathcal{O}(T^{-1}\log T)\\
    &\leq \sqrt{8T^{-1}\log T}\frac{\Sigma_{2}-2\sigma^{2}_{\min}}{\sqrt{\alpha_{K}}}+\sqrt{8T^{-1}\log T}\sum_{k=1}^{K} \frac{\sigma^{2}_{k}}{\sqrt{\alpha_{k}}}+\mathcal{O}(T^{-1}\log T)\\
    &\leq \sqrt{8T^{-1}\log T}\left[\frac{\sqrt{\Sigma_{2}}(\Sigma_{2}-2\sigma^{2}_{\min})}{\sigma_{\min}}+\sqrt{\Sigma_{2}}\Sigma_{1}\right]+\mathcal{O}(T^{-1}\log T)
    \end{align*}
Then we can get that:
\begin{align*}
     \mathbb{E}\left[R_{\infty}(\bm{n}_{\pi_2}) - R_{\infty}(\bm{n}^{*})\right] &= 
    \mathbb{E}\left[R_{\infty}(\bm{\lambda}) - R_{\infty}(\bm{\lambda}^{*})\right] \\
    &\leq \mathbb{E}\left[R_{\infty}(\bm{\lambda}) - R_{\infty}(\bm{\lambda}^{*})|\xi_{T}(\delta)\right]+ 2T^{-1}\mathbb{E}\left[R_{\infty}(\bm\lambda) - R_{\infty}(\bm{\lambda}^{*})|\xi^{c}_{T}(\delta)\right] \\
    &\leq 2\sqrt{2}\left[\frac{\sqrt{\Sigma_{2}}(\Sigma_{2}-2\sigma^{2}_{\min})}{\sigma_{\min}}+\sqrt{\Sigma_{2}}\Sigma_{1}\right]T^{-3/2}\sqrt{\log T}+o(T^{-3/2})
\end{align*}
When $p$ is finite, in the event $\xi_{T}(\delta)$, we would have:
\begin{align*}
    \sum_{k=1}^{K}\frac{(\lambda_{k,\pi_2}-\lambda^{*}_{k})^{2}}{\lambda^{*}_{k}} &\leq \frac{\Sigma_{q}}{(\sum_{k=1}^{K}(\sigma^{2}_{k}-\varepsilon^{-}_{\tau_{k}})^{q/2})^{2}}\sum_{k=1}^{K}\frac{[(\sigma^{2}_{k}+\varepsilon^{+}_{\tau_{k}})^{q/2}-\sigma^{q}_{k}]^{2}}{\sigma_{k}^{q}} \\
    &= \frac{\sum_{k=1}^{K}\sigma^{q}_{k}[(1+s^{+}_{\tau_{k}})^{q/2}-1]^{2}}{[\sum_{k=1}^{K}\sigma^{q}_{k}(1-s^{-}_{\tau_{k}})^{q}]^{2}} \Sigma_{q}\\
    &\leq \frac{q^{2}\sum_{k=1}^{K}\sigma^{q}_{k}(s^{+}_{\tau_{k}})^{2}}{4\Sigma_{q}(1-s^{-}_{\tau_{\min}})^{2q}} \\
    &\leq \frac{5q^{2}\sum_{k=1}^{K}\sigma^{q}_{k}/\alpha_{k}}{2\Sigma_{q}T}\log T+o(T^{-1})\\
    &\leq \frac{5}{2}Kq^{2}T^{-1}\log T+o(T^{-1})
\end{align*}
Then we have:
\begin{align*}
    \mathbb{E}\left[R_{\infty}(\bm{n}_{\pi_2}) - R_{\infty}(\bm{n}^{*})\right] &= 
    \mathbb{E}\left[R_{\infty}(\bm{\lambda}) - R_{\infty}(\bm{\lambda}^{*})\right] \\
    &\leq \mathbb{E}\left[R_{\infty}(\bm{\lambda}) - R_{\infty}(\bm{\lambda}^{*})|\xi_{T}(\delta)\right]+ 2T^{-3/2}\mathbb{E}\left[R_{\infty}(\bm{\lambda}) - R_{\infty}(\bm{\lambda}^{*})|\xi^{c}_{T}(\delta)\right] \\
    &\leq  \frac{5}{4}(p+1)Kq^{2}(\Sigma_{q})^{2/q}T^{-2}\log T+o(T^{-2}) \\
    &= \frac{5Kp^{2}(\Sigma_{q})^{2/q}}{p+1}T^{-2}\log T+o(T^{-2})
\end{align*}    
\end{proof}

\subsection{Proof of Theorem~\ref{thm:Contextual_regret_SSG}}
\begin{proof}
   This result is directly from Lemma~\ref{lemma:MSE_bound} and Theorem~\ref{thm:regretAlg3_strictlysubgaussian}.   
\end{proof}

\end{document}